%% file: main.tex
\pdfoutput=1
\PassOptionsToPackage{table}{xcolor}
\documentclass[runningheads]{llncs}

\usepackage{eccv}
\usepackage{placeins}
\usepackage{eccvabbrv}
\usepackage{graphicx}
\usepackage{booktabs}
\usepackage{xcolor}
\usepackage{iftex}
\usepackage{hyperref}
\usepackage{orcidlink}
\hypersetup{hidelinks}

\definecolor{AccelAesBlue}{HTML}{1F77B4}
\definecolor{AccelAesRow}{HTML}{EAF3FF}
\definecolor{AccelAesGreen}{HTML}{DDF2E6}
\newcommand{\oursrow}{\rowcolor{AccelAesRow}}
\newcommand{\goodcell}[1]{\cellcolor{AccelAesGreen}\textbf{#1}}
\newcommand{\arxivonly}[1]{#1}

\begin{document}

\title{AccelAes: Accelerating Diffusion Transformers for Training-Free Aesthetic-Enhanced Image Generation}
\titlerunning{AccelAes: Training-Free Aesthetic-Enhanced DiT Acceleration}

\author{Xuanhua Yin \and
Chuanzhi Xu \and
Haoxian Zhou \and
Boyu Wei \and
Weidong Cai\thanks{Corresponding author.}}

\authorrunning{X. Yin et al.}

\institute{School of Computer Science, The University of Sydney, NSW 2006, Australia
\email{\{xuanhua.yin,chuanzhi.xu,hzho0442,bwei0951,tom.cai\}@sydney.edu.au}}

\maketitle

\begin{abstract}
Diffusion Transformers (DiTs) are a dominant backbone for high-fidelity text-to-image generation due to strong scalability and alignment at high resolutions. However, quadratic self-attention over dense spatial tokens leads to high inference latency and limits deployment. We observe that denoising is spatially non-uniform with respect to aesthetic descriptors in the prompt. Regions associated with aesthetic tokens receive concentrated cross-attention and show larger temporal variation, while low-affinity regions evolve smoothly with redundant computation. Based on this insight, we propose \textbf{AccelAes}, a training-free framework that accelerates DiTs through aesthetics-aware spatio-temporal reduction while improving perceptual aesthetics. AccelAes builds AesMask, a one-shot aesthetic focus mask derived from prompt semantics and cross-attention signals. When localized computation is feasible, SkipSparse reallocates computation and guidance to masked regions. We further reduce temporal redundancy using a lightweight step-level prediction cache that periodically replaces full Transformer evaluations. Experiments on representative DiT families show consistent acceleration and improved aesthetics-oriented quality. On Lumina-Next, AccelAes achieves a \textbf{2.11$\times$} speedup and improves ImageReward by \textbf{+11.9\%} over the dense baseline. Code is available at \href{https://github.com/xuanhuayin/AccelAes}{https://github.com/xuanhuayin/AccelAes}.
\end{abstract}

\section{Introduction}
\input{chapters/Introduction.tex}

\section{Related Work}
\input{chapters/Related.tex}

\section{Methodology}
\input{chapters/Methodology.tex}

\section{Experiments and Results}
\input{chapters/Experiments.tex}

\section{Conclusion}
\input{chapters/Conclusion}

\FloatBarrier
\bibliographystyle{splncs04}
\bibliography{main}

\clearpage
\appendix
\section*{Appendix}
\input{chapters/Appendix}

\end{document}

%% file: chapters/Introduction.tex
Diffusion models have become dominant for high-fidelity text-to-image generation~\cite{1,4}. Recent research increasingly adopts Transformer backbones, namely the Diffusion Transformer (DiT)~\cite{9,10,11,12}, to achieve better scalability and improved visual quality. Compared with earlier U-Net based pipelines~\cite{4}, DiTs provide higher capacity and stronger alignment at high resolutions~\cite{9}, enabling the generation of more detailed, realistic, and semantically consistent images.

\begin{figure*}[t]
  \centering
  \includegraphics[width=0.95\linewidth]{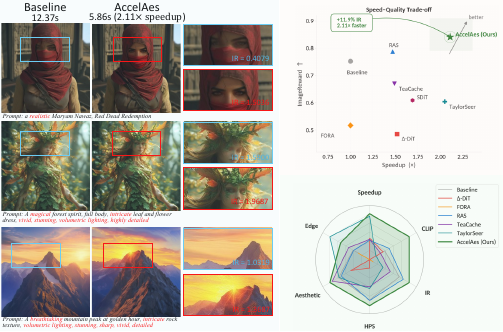}
\caption{\textbf{AccelAes improves both efficiency and overall quality.}
Left: Qualitative comparison under identical prompts, with zoomed-in crops and per-sample ImageReward (IR).
Right: Speed--quality summary, and a bottom-right multi-metric radar on \textbf{Lumina} benchmarking AccelAes against the dense baseline and representative training-free acceleration methods, demonstrating consistent gains across metrics.}
  \label{fig:teaser}
\end{figure*}
However, the performance comes with a more immediate cost: inference becomes significantly slower. At each denoising step, a DiT must perform full self-attention over all spatial tokens, and the pairwise interactions among tokens make every forward pass computationally expensive. Moreover, high-resolution image generation typically requires dozens of denoising steps, leading to cumulative inference latency across iterations. In practice, model performance is often constrained not only by model size but also by repeated computation and runtime overhead during sampling. As attention operations and memory movement account for an increasing portion of total execution time, it has become a central challenge for DiTs to reduce redundant computation at inference while maintaining visual quality ~\cite{7}. This motivates our focus on training-free acceleration that explicitly accounts for perceptual and aesthetics-oriented criteria.

To address this issue, many works aim to accelerate diffusion inference without retraining. Some research reduces the total number of denoising steps through improved samplers~\cite{13,14,20,21,22}. Others reduce per-step computation, for example, through token merging~\cite{26,27,28,29,30}, structured sparse attention~\cite{30}, or intermediate feature caching~\cite{13,14,15,16,18,20,21,22,23,26}. However, these methods share a common limitation: they apply uniform strategies across the spatial dimension, either merging or pruning all tokens in the same manner, or caching and skipping computation uniformly over the entire image. Such strategies are often vulnerable in practice because image generation is not spatially uniform. Certain regions, such as main subjects, fine textures, and lighting boundaries, have a much greater impact on perceptual quality, while other regions, such as smooth backgrounds, contribute less to the final visual impression \cite{104,105}. If computation is aggressively reduced in perceptually important regions, image quality degrades significantly. If conservative settings are applied everywhere to protect quality, much of the potential speedup is lost. This leads to a natural question: can computation be allocated adaptively across space according to content importance?

From another perspective, text-to-image models based on DiTs are no longer solely optimized for strict semantic alignment with the prompt, but are increasingly designed to generate images that better reflect human preferences and aesthetic judgments~\cite{36,37,38,39,42}. With advances in alignment methods and human feedback optimization methods such as RLHF~\cite{101,100}, modern generative models place greater emphasis on aesthetic attributes, including ``intricate'', ``volumetric-lighting'', and ``detailed''~\cite{34,35}. These descriptors often appear as adjectives or stylistic tokens in prompts and influence image generation via cross-attention mechanisms~\cite{4,103}. In other words, the model is not only concerned with what to generate, but also with how to generate it well.

This trend suggests an important insight: different spatial regions contribute unequally to aesthetic quality. Regions containing main subjects with fine textures, lighting transitions, and material details often dominate human perception and aesthetic judgment, while large smooth background areas typically have a smaller influence on aesthetic scores~\cite{34, 104}. Moreover, cross-attention explicitly models the correspondence between text tokens and image patches in modern text-to-image diffusion models\cite{4, 103}. In particular, tokens associated with aesthetic descriptors tend to induce highly concentrated attention patterns over specific spatial regions. This suggests that during denoising, the model implicitly differentiates between aesthetically sensitive regions and less sensitive regions.

However, existing DiT acceleration methods rarely exploit this signal. They typically apply uniform token reduction based on numerical similarity or fixed structural rules, overlooking the spatial non-uniformity that is closely tied to human preference. In a generation paradigm that increasingly emphasizes aesthetic quality, such semantically unaware compression is suboptimal. 

We find that spatial aesthetic non-uniformity and temporal redundancy are coupled. Regions related to aesthetic descriptors vary more across timesteps and are harder to extrapolate, while low-affinity regions evolve smoothly and are more reusable. This implies that the attention structure already indicates which regions require precise computation and which can be skipped. Leveraging both signals enables content-aware acceleration while preserving key aesthetic quality.

Based on the above observations, we propose \textbf{AccelAes (Accelerated Aesthetics)}, a training-free acceleration framework for DiTs. The core idea of AccelAes is to leverage aesthetics-driven spatial non-uniformity and temporal prediction redundancy during the denoising process, allocating computation preferentially to perceptually sensitive regions while compressing redundant computation in less critical areas. Specifically, we use internal attention signals to construct AesMask, an aesthetics-aware semantic mask that identifies spatial regions associated with aesthetic descriptors, and refine them more carefully when the architecture permits localized updates. Meanwhile, we introduce SkipSparse, a unified acceleration module that combines spatially-adaptive sparse computation with an output-level step prediction reuse mechanism (StepCache) along the temporal dimension to reduce redundant computation across timesteps. These two mechanisms complement each other, enabling effective inference acceleration while preserving key perceptual quality. Fig.~\ref{fig:teaser} previews this speed--quality tension and shows that AccelAes improves both efficiency and overall quality on representative prompts and metrics.
Our contributions are summarized as follows:

\begin{itemize}
\item We propose \textbf{AccelAes}, a training-free acceleration framework for Diffusion Transformers, motivated by aesthetics-driven spatial non-uniformity and temporal redundancy during denoising.

\item We introduce an aesthetics-aware acceleration design that couples \textbf{AesMask} and \textbf{SkipSparse} to localize perceptually sensitive regions from internal model signals and reallocate computation and guidance accordingly, while compressing less critical updates.

\item We conduct extensive experiments on three representative DiT backbones, showing that AccelAes substantially reduces inference time while consistently improving aesthetics- and preference-oriented quality, yielding a markedly better speed--quality trade-off.
\end{itemize}

%% file: chapters/Related.tex
\subsection{Acceleration of DiT for Image Generation}
Diffusion-based generators synthesize images via iterative denoising, making sampling computationally intensive and motivating training-free acceleration that improves efficiency without modifying backbone parameters~\cite{1,2,4}. Prior work mainly follows two directions: reducing sampling steps or changing the sampling trajectory with improved solvers and controlled exploration~\cite{3,6,ctrlzsampling}, or reducing per-step cost by exploiting temporal and spatial redundancy, e.g., feature caching/reuse~\cite{13,16,20,22,23}, timestep forecasting/skipping~\cite{15,21}, and token-level compression (pruning/merging) or training-free sparse attention variants~\cite{19,26,27,28,29,30}.

Many methods decide reductions using numerical similarity, heuristics, or generic saliency proxies rather than perceptual objectives, so aggressive compression can remove compute from visually important regions and introduce localized artifacts. Region-adaptive methods such as RAS and SDiT allocate non-uniform budgets across spatial tokens~\cite{31,32}, but still rely on proxy signals (e.g., attention concentration, heuristic complexity, or semantic grouping) to define region priorities~\cite{31,32}. Meanwhile, modern text-to-image systems increasingly optimize for human preference and aesthetic quality beyond semantic correctness~\cite{100,101,36}, yet such image-level objectives do not directly yield stable region priorities for inference-time budgeting. Moreover, trajectory-level acceleration (e.g., caching/forecasting) is typically applied uniformly over space, leaving open where computation should be preserved under limited budgets.

\subsection{Aesthetic and Guidance Signals for Image Generation}
Inference-time guidance offers a practical control interface for text-to-image diffusion models: CLIP supports text--image semantic consistency~\cite{8}, and classifier-free guidance (CFG) strengthens prompt adherence by scaling the conditional--unconditional gap~\cite{5}. However, a single global guidance scale can be spatially inconsistent and induce local semantic/structural issues, motivating spatially varying guidance that applies different strengths to semantic regions without additional training~\cite{35}; related work further enables continuous modulation of attribute intensity~\cite{37}. These approaches keep dense computation and do not address how guidance should interact with inference-time compute budgeting, whereas we explicitly couple guidance with spatial compute allocation in a training-free acceleration setting.

Human preference and aesthetic assessment provides complementary signals for generative models. Preference datasets and reward models such as Pick-a-Pic and ImageReward align objectives with human feedback~\cite{33,34}, and preference feedback can be used for post-training improvements such as step-by-step preference optimization~\cite{36}. For evaluation, preference and aesthetics are often measured by image-level scoring functions, including CLIP-based aesthetic predictors~\cite{42}, benchmarks such as HPSv2~\cite{41}, and classic datasets like AVA~\cite{104}; aesthetic-related rewards have also been used for automatic prompt optimization~\cite{40}. In our work, preference-related models are used for evaluation rather than retraining, but their scalar image-level scores are difficult to translate into region-level priorities or token-level compute budgets. This leaves an interface problem for training-free acceleration: how to turn aesthetic objectives into executable spatial budgeting rules. We address this gap by extracting aesthetic-related cues from internal attention patterns conditioned on aesthetic descriptors and using them to guide spatial compute allocation while preserving global context.

%% file: chapters/Methodology.tex
We propose \textbf{AccelAes}, a training-free acceleration framework for Diffusion Transformers (DiTs) based on aesthetics-driven spatio-temporal non-uniformity. Given a text prompt, AccelAes first constructs \textbf{AesMask}, an aesthetics-aware semantic mask over image tokens from cross-attention signals. It then applies \textbf{SkipSparse}, which performs spatially-adaptive sparse computation when localized updates are supported. SkipSparse also includes a universal step-level prediction cache that reuses noise predictions across timesteps. Together, AesMask decides \emph{where} to spend compute, and SkipSparse decides \emph{how} to reduce redundant computation in both space and time.

\begin{figure*}[t]
  \centering
  \includegraphics[width=0.95\linewidth]{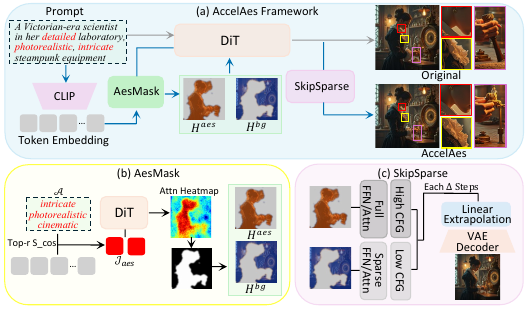}
  \caption{\textbf{The proposed AccelAes framework.}
(a) AccelAes builds AesMask from prompt semantics and cross-attention. It uses the mask for spatially adaptive computation and region-wise guidance. SkipSparse also includes a step-level cache with linear extrapolation to reduce redundant forwards across timesteps.
(b) AesMask uses aesthetic anchors, CLIP similarity, and cross-attention aggregation, then applies percentile thresholding to obtain a binary mask.
(c) SkipSparse updates attention and FFN mainly on aesthetic-focus tokens. It keeps global context with global keys/values and reuses predictions across steps via the cache.}
  \label{fig:pipeline}
\end{figure*}
\subsection{Preliminaries: Aesthetically-Driven Non-Uniformity}
We consider diffusion sampling in latent space. Let $\mathbf{z}_t \in \mathbb{R}^{d \times N}$ be latent tokens at timestep $t$, where $N$ is the number of spatial tokens. A DiT parameterized by $\theta$ predicts noise (or velocity) as: 
\begin{equation}
\hat{\boldsymbol{\epsilon}}_t = f_{\theta}\!\left(\mathbf{z}_t, \mathbf{c}, t\right),
\label{eq:eps_pred}
\end{equation}
where $\mathbf{c}$ is the text condition and $\hat{\boldsymbol{\epsilon}}_t \in \mathbb{R}^{d \times N}$ is the per-token prediction.

DiT blocks use self-attention and cross-attention between image tokens and text tokens. Let $\mathbf{y} = [\mathbf{y}_1,\ldots,\mathbf{y}_M] \in \mathbb{R}^{d_c \times M}$ be $M$ text token embeddings. For a cross-attention layer $\ell$ at timestep $t$, we denote its weights by: 
\begin{equation}
\mathbf{A}^{(\ell)}_t \in [0,1]^{N \times M}, 
\qquad
\sum_{j=1}^{M} \mathbf{A}^{(\ell)}_t[i,j]=1 \ \ \forall i,
\label{eq:cross_attn}
\end{equation}
where $\mathbf{A}^{(\ell)}_t[i,j]$ measures how much image token $i$ attends to text token $j$.

We observe a consistent non-uniformity for aesthetics-oriented prompts. Attention mass concentrates on a subset of spatial tokens linked to aesthetic descriptors. These tokens often control textures, highlights, and boundaries that dominate perceptual judgment. Other tokens are less sensitive and change more smoothly across nearby timesteps. AccelAes uses this property to reduce computation where it is less critical and to preserve computation where it matters.

\subsection{AesMask: Aesthetics-Aware Semantic Mask Generation}
We build a binary mask over image tokens for \emph{aesthetic-focus regions}. This module is \textbf{AesMask}, as shown in Fig.~\ref{fig:pipeline} (b). In text-conditioned DiTs, the prompt affects image tokens through cross-attention. We use this internal signal to locate tokens related to aesthetic-relevant prompt semantics, and we mark them for later selective updates.

We define a set of aesthetic anchor words:
\begin{equation}
\mathcal{A}=\{a_1,\ldots,a_K\},
\label{eq:anchors}
\end{equation}
where each $a_k$ is a textual descriptor (e.g., ``cinematic'', ``intricate''). The anchors provide a compact reference that helps us pick aesthetic-related prompt tokens without external supervision.
\arxivonly{The complete anchor vocabulary and the default AesMask hyperparameters are provided in Appendix~\ref{app:anchors} and Table~\ref{tab:3}.}

\paragraph{Aesthetic Prompt Tokens.}
Let $e(\cdot)$ be a frozen text encoder (CLIP text encoder). For each prompt token $\mathbf{y}_j$, we compute cosine similarity to each anchor $a_k$:
\begin{equation}
s_{j,k} = \frac{\langle e(\mathbf{y}_j), e(a_k)\rangle}{\|e(\mathbf{y}_j)\|\ \|e(a_k)\|}.
\label{eq:clip_sim}
\end{equation}
We assign each token an aesthetic score:
\begin{equation}
\alpha_j = \max_{k\in\{1,\ldots,K\}} s_{j,k},
\label{eq:alpha_j}
\end{equation}
and select the top-$r$ tokens by $\alpha_j$. Their index set is $\mathcal{J}_{\text{aes}} \subseteq \{1,\ldots,M\}$. These tokens represent the aesthetic-related part of the prompt that we will use when aggregating attention.

\paragraph{Cross-attention Aggregation.}
At timestep $t$, each cross-attention layer $\ell$ outputs $\mathbf{A}^{(\ell)}_t \in \mathbb{R}^{N\times M}$. We aggregate weights for the selected aesthetic tokens:
\begin{equation}
m_t[i] = \frac{1}{|\mathcal{L}|}\sum_{\ell \in \mathcal{L}}\ \sum_{j \in \mathcal{J}_{\text{aes}}}\mathbf{A}^{(\ell)}_t[i,j],
\label{eq:affinity_map}
\end{equation}
where $m_t \in \mathbb{R}^{N}$ is an affinity map over image tokens and $\mathcal{L}$ is the set of layers used. A larger $m_t[i]$ means token $i$ consistently attends to aesthetic-relevant text tokens. We treat it as an internal proxy for perceptual importance.

\paragraph{Percentile Thresholding.}
We binarize $m_t$ with a percentile threshold:
\begin{equation}
\mathbf{M}_t[i] = \mathbb{I}\!\left(m_t[i] \ge \mathrm{Perc}\!\left(m_t,\,p\right)\right),
\label{eq:mask}
\end{equation}
where $\mathbf{M}_t \in \{0,1\}^{N}$ and $p\in(0,100)$ controls sparsity. We define $\mathcal{I}_t=\{i \mid \mathbf{M}_t[i]=1\}$. The percentile form adapts to the scale of $m_t$ across prompts, so it avoids tuning an absolute threshold. AesMask is training-free and uses cross-attention and a frozen text encoder. Its output $\mathbf{M}_t$ (or $\mathcal{I}_t$) is passed to SkipSparse.

\subsection{Spatially-Adaptive Sparse Computation}
When localized updates are supported, we use $\mathbf{M}_t$ to reduce computation. This module is \textbf{SkipSparse}, as shown in Fig.~\ref{fig:pipeline} (c). SkipSparse focuses updates on tokens in $\mathcal{I}_t$ and keeps global context through global keys/values. This is important because it allows focus tokens to read information from the full image.

Let $\mathbf{H}_t \in \mathbb{R}^{d \times N}$ be the hidden states entering a Transformer block at timestep $t$. We split tokens into focus and background:
\begin{equation}
\mathbf{H}_t = \left[\mathbf{H}^{\text{aes}}_t,\ \mathbf{H}^{\text{bg}}_t\right],
\label{eq:partition}
\end{equation}
where $\mathbf{H}^{\text{aes}}_t$ contains tokens in $\mathcal{I}_t$ and $\mathbf{H}^{\text{bg}}_t$ contains the rest. We scatter updated tokens back to the original order after each block so the block output matches the dense baseline shape.

\paragraph{Self-attention with Global Keys/Values.}
We update attention only for focus queries and keep keys/values global:
\begin{equation}
\mathrm{Attn}\!\left(\mathbf{Q}^{\text{aes}}_t,\mathbf{K}^{\text{all}}_t,\mathbf{V}^{\text{all}}_t\right)
=
\mathrm{Softmax}\!\left(\frac{\left(\mathbf{Q}^{\text{aes}}_t\right)^{\top}\mathbf{K}^{\text{all}}_t}{\sqrt{d_h}}\right)\mathbf{V}^{\text{all}}_t,
\label{eq:qkv_sparse}
\end{equation}
where $\mathbf{Q}^{\text{aes}}_t = \mathbf{W}_Q \mathbf{H}^{\text{aes}}_t$, $\mathbf{K}^{\text{all}}_t = \mathbf{W}_K \mathbf{H}_t$, and $\mathbf{V}^{\text{all}}_t = \mathbf{W}_V \mathbf{H}_t$. This reduces the number of queries from $N$ to $|\mathcal{I}_t|$ and keeps long-range interactions through global keys/values.

\paragraph{FFN on Aesthetic Tokens.}
We apply FFN only to focus tokens and reuse a cached background activation:
\begin{equation}
\tilde{\mathbf{H}}^{\text{aes}}_t = \mathrm{FFN}\!\left(\mathbf{H}^{\text{aes}}_t\right), 
\qquad
\tilde{\mathbf{H}}^{\text{bg}}_t = \mathbf{C}^{\text{bg}}_t,
\label{eq:ffn_cache}
\end{equation}
where $\mathbf{C}^{\text{bg}}_t$ is taken from the most recent full update. This avoids recomputing FFN on background tokens and still allows refinement on aesthetic-focus tokens.

\paragraph{Spatial CFG.}
Let $\hat{\boldsymbol{\epsilon}}^{\text{cond}}_t$ and $\hat{\boldsymbol{\epsilon}}^{\text{uncond}}_t$ be conditional and unconditional predictions. We apply a spatially varying guidance scale:
\begin{equation}
\hat{\boldsymbol{\epsilon}}^{\text{cfg}}_t[i] =
\hat{\boldsymbol{\epsilon}}^{\text{uncond}}_t[i]
+
g_t[i]\left(\hat{\boldsymbol{\epsilon}}^{\text{cond}}_t[i]-\hat{\boldsymbol{\epsilon}}^{\text{uncond}}_t[i]\right),
\label{eq:spatial_cfg}
\end{equation}
where $g_t[i] = g_{\text{bg}} + (g_{\text{aes}}-g_{\text{bg}})\mathbf{M}_t[i]$ and $g_{\text{aes}}\ge g_{\text{bg}}\ge 1$. This places stronger guidance on focus tokens and keeps background guidance milder.

\subsection{Universal Step-Level Prediction Caching}
SkipSparse also includes a universal step-level cache, as shown in Fig.~\ref{fig:pipeline} (c). It reuses the final output $\hat{\boldsymbol{\epsilon}}_t$ and works across DiT variants. It targets a simple fact that nearby timesteps often have correlated predictions.

We cache predictions at selected timesteps and reuse them with linear extrapolation. Let $\Delta$ be the stride between full forwards. Let $\tau$ be the latest timestep with a full forward. We store: 
\begin{equation}
\mathbf{E}_{\tau} = \hat{\boldsymbol{\epsilon}}_{\tau}, 
\qquad
\mathbf{E}_{\tau+\Delta} = \hat{\boldsymbol{\epsilon}}_{\tau+\Delta}.
\label{eq:cache_store}
\end{equation}
For an intermediate timestep $t\in(\tau,\tau+\Delta)$, we approximate: 
\begin{equation}
\hat{\boldsymbol{\epsilon}}_{t} \approx \mathbf{E}_{\tau} + \lambda(t)\left(\mathbf{E}_{\tau}-\mathbf{E}_{\tau+\Delta}\right),
\label{eq:linear_extrap}
\end{equation}
where $\lambda(t)=\frac{t-\tau}{\Delta}$. We use a warmup of $T_w$ initial steps with full inference. After warmup, we refresh the cache every $\Delta$ steps by running a full forward and updating Eq.~\ref{eq:cache_store}. This periodic refresh limits drift and keeps stable.

%% file: chapters/Experiments.tex

\subsection{Experimental Setup}
\label{sec:exp_setup}
\begin{table*}[t]
\centering
\small
\setlength{\tabcolsep}{4pt}
\renewcommand{\arraystretch}{1.15}
\caption{\textbf{Main results on three DiT backbones.}
We report runtime and generation-quality metrics for AccelAes and competing methods on Lumina-Next, SD3-Medium, and FLUX.1-dev, showing consistent speedup with improved overall quality.}
\label{tab:1}
\resizebox{\linewidth}{!}{%
\begin{tabular}{l l cc c c c c c}
\toprule
Models & Methods & Time$\downarrow$ & Speedup$\uparrow$ & CLIP$\uparrow$ & IR$\uparrow$ & HPS$\uparrow$ & Aesth$\uparrow$ & Edge$\uparrow$ \\
\midrule
Lumina & Baseline & 12.37 & 1.00$\times$ & 0.2531 & 0.7518 & 0.2710 & 5.9407 & 0.5829 \\
\oursrow Lumina & \textbf{AccelAes} & \textbf{5.86} & \textbf{2.11$\times$} & \textbf{0.2640} & \textbf{0.8410} & \textbf{0.2740} & \textbf{6.0414} & \textbf{0.6285} \\
\midrule
SD3 & Baseline & 3.52 & 1.00$\times$ & 0.2662 & 0.8788 & 0.2895 & 5.7330 & 0.9160 \\
\oursrow SD3 & \textbf{AccelAes} & \textbf{2.34} & \textbf{1.50$\times$} & \textbf{0.2744} & \textbf{0.9042} & \textbf{0.3014} & \textbf{5.9898} & \textbf{0.9437} \\
\midrule
FLUX & Baseline & 12.66 & 1.00$\times$ & 0.2753 & 1.233 & 0.3200 & 6.243 & 0.560 \\
\oursrow FLUX & \textbf{AccelAes} & \textbf{7.31} & \textbf{1.73$\times$} & \textbf{0.2772} & \textbf{1.317} & \textbf{0.3214} & \textbf{6.372} & \textbf{0.590} \\
\bottomrule
\end{tabular}
}
\end{table*}

\paragraph{Models and Baselines.}
We evaluate AccelAes on three Diffusion Transformer backbones: Lumina-Next-T2I~\cite{11}, SD3-Medium~\cite{43}, and FLUX.1-dev~\cite{44}. We follow the standard DiT formulation~\cite{9} and only modify inference-time computation.
For training-free acceleration baselines, we include $\Delta$-DiT~\cite{14}, FORA~\cite{15}, Region-Adaptive Sampling (RAS)~\cite{31}, TeaCache~\cite{17}, and TaylorSeer~\cite{21}. For each backbone, all methods are evaluated under the same backbone-specific sampler and guidance configuration, and we adhere to the same implementation-level rules when reproducing baselines.

\paragraph{Evaluation Protocol.}
We use a unified protocol with 10{,}000 prompts from Pick-a-Pic and three fixed seeds per prompt (seeds $\{1,2,3\}$; 30{,}000 images per configuration) for both main results and ablations. Within each table, all AccelAes variants and baselines share the same prompt subset and seeds to enable paired comparisons. All methods are implemented in a unified PyTorch pipeline without architecture-specific engines, kernel-level optimizations, or external accelerators, and we follow backbone-default sampling settings. We evaluate all backbones at $1024\times1024$ resolution using 30 steps for Lumina-Next-T2I and 28 steps for SD3-Medium and FLUX.1-dev. Runtime is reported as end-to-end per-image latency from the start of sampling to the final decoded image, including VAE decoding and the two-pass CFG forward when enabled. Unless specified otherwise, AccelAes uses a cache refresh interval of $\Delta=2$ with warmup $T_w=5$, a one-shot AesMask at \texttt{mask\_step=5}, and the same fixed 34-anchor vocabulary across paired comparisons.
\arxivonly{Backbone-specific cache settings and the full evaluation environment are detailed in Appendix~\ref{app:stepcache} and Appendix~\ref{app:runtime}.}
\begin{table}[!tbp]
\centering
\small
\setlength{\tabcolsep}{4pt}
\renewcommand{\arraystretch}{1.10}
\caption{\textbf{Lumina-Next-T2I: baseline comparisons.}
Comparison of AccelAes with representative acceleration baselines on Lumina-Next-T2I using the main metric set.}
\label{tab:2}
\resizebox{\linewidth}{!}{%
\begin{tabular}{l cc c c c c c}
\toprule
Methods & Time$\downarrow$ & Speedup$\uparrow$ & CLIP$\uparrow$ & IR$\uparrow$ & HPS$\uparrow$ & Aesth$\uparrow$ & Edge$\uparrow$ \\
\midrule
Lumina-Next-T2I~\cite{11} & 12.37 & 1.00$\times$ & 0.2531 & 0.752 & 0.2710 & 5.941 & 0.583 \\
$\Delta$-DiT~\cite{14} & 8.13 & 1.52$\times$ & 0.2523 & 0.485 & 0.2540 & 5.873 & 0.522 \\
FORA~\cite{15} & 12.36 & 1.00$\times$ & 0.2463 & 0.517 & 0.2496 & 5.766 & 0.564 \\
RAS~\cite{31} & 8.38 & 1.47$\times$ & 0.2550 & 0.788 & 0.2713 & 5.927 & 0.581 \\
SDiT~\cite{32} & 7.30 & 1.69$\times$ & 0.2502 & 0.6093 & 0.2538 & 5.822 & 0.4289 \\
TeaCache~\cite{17} & 8.28 & 1.49$\times$ & 0.2512 & 0.670 & 0.2640 & 5.978 & 0.599 \\
TaylorSeer~\cite{21} & 6.02 & 2.05$\times$ & 0.2491 & 0.604 & 0.2650 & 5.940 & \textbf{0.668} \\
\oursrow \textbf{AccelAes (ours)} & \textbf{5.86} & \textbf{2.11$\times$} & \textbf{0.2640} & \textbf{0.841} & \textbf{0.2740} & \textbf{6.041} & 0.629 \\
\bottomrule
\end{tabular}
}
\end{table}

\begin{figure*}[tbp]
  \centering
  \includegraphics[width=\linewidth]{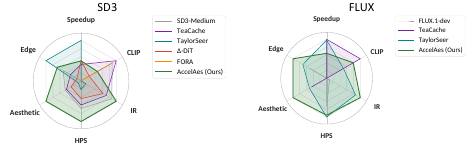}
\caption{\textbf{Baseline comparisons on SD3 and FLUX.}
We compare AccelAes with the dense baseline and prior acceleration methods on SD3 and FLUX using the same prompts, seeds, and measurement protocol. AccelAes achieves a better speed quality trade off and improves aesthetics oriented scores while reducing sampling time.}
  \label{fig:3}
\end{figure*}

\paragraph{Metrics and Measurement.}
We report runtime and six automatic metrics: Time (seconds per image) and Speedup defined as $\mathrm{Time}_{\mathrm{baseline}}/\mathrm{Time}_{\mathrm{method}}$, CLIP score (CLIP)~\cite{8} for text--image alignment, ImageReward (IR)~\cite{34} and HPSv2 (HPS)~\cite{41} as preference-oriented proxies, LAION Aesthetic Score (Aesth) computed with the LAION aesthetic predictor~\cite{42}, and Edge Density (Edge) as a lightweight sharpness/detail proxy. Time is averaged over the full evaluation set and includes the complete sampling trajectory, including conditional/unconditional passes for CFG. Unless otherwise specified, AccelAes constructs a one-shot semantic mask at \texttt{mask\_step=5} and reuses it for the entire sampling trajectory and classifier-free guidance follows the standard formulation~\cite{5}.




\subsection{Main Results}
\label{sec:main_results}

\paragraph{Overall Results across Backbones.}
Table~\ref{tab:1} shows that AccelAes improves the speed--quality trade-off across three DiT backbones.
The key pattern is that acceleration does not come from uniformly weakening denoising.
Instead, AccelAes reduces compute where the denoising trajectory is redundant and preserves updates where aesthetic cues are concentrated.
This leads to simultaneous runtime reduction and stronger preference-oriented metrics.
On Lumina-Next, AccelAes reaches the highest speedup and also yields a clear improvement on ImageReward together with gains on CLIP, HPS, Aesth, and Edge.
On SD3-Medium, AccelAes again improves ImageReward and HPS under a 1.50$\times$ speedup.
On FLUX.1-dev, AccelAes maintains a strong balance, which suggests the same mechanism transfers beyond a single architecture and sampler.
Overall, these results support that aesthetics-driven spatio-temporal non-uniformity is a reliable signal for compute reduction under matched prompts, seeds, and backbones.
\arxivonly{The full SD3-Medium and FLUX.1-dev numerical comparisons are reported in Appendix~\ref{app:sd3} and Appendix~\ref{app:flux}.}

\subsection{Human and Robustness Validation}
\label{sec:robustness_validation}

\begin{table*}[!tbp]
\centering
\small
\setlength{\tabcolsep}{6pt}
\renewcommand{\arraystretch}{1.12}
\caption{\textbf{Blind pairwise human preference study.}
Win rates exclude ties. Each comparison uses randomized left/right ordering and four questions: overall preference, prompt alignment, visual appeal, and fewer artifacts.}
\label{tab:rebuttal_validation}
\begin{tabular}{lcccc}
\toprule
Comparison & Overall & Alignment & Appeal & Fewer artifacts \\
\midrule
AccelAes vs.\ Dense & 67\% & 59\% & 68\% & 64\% \\
\oursrow AccelAes vs.\ FORA & \goodcell{89\%} & \goodcell{90\%} & \goodcell{87\%} & \goodcell{94\%} \\
AccelAes vs.\ w/o Spatial-CFG & 66\% & 69\% & 66\% & 68\% \\
\bottomrule
\end{tabular}
\end{table*}

Table~\ref{tab:rebuttal_validation} adds a direct human validation of the automatic metric gains.
The study uses 15 annotators, 60 pairs, and four questions per pair.
AccelAes is preferred over the dense baseline, FORA, and the variant without Spatial CFG across all criteria.
Binomial tests on non-tie votes reject the 50\% chance null for the main preference comparisons ($p<0.001$, except alignment against the dense baseline).
We further checked whether these gains depend on prompt wording or simple scene bias.
On descriptor-free prompts, AccelAes improves ImageReward from 1.109 to 1.144 (+3.1\%); with aesthetic descriptors, it improves the stronger dense baseline from 1.168 to 1.224 (+4.8\%).
Across six hard-scene categories, the ImageReward gain remains positive for single-subject (+1.8\%), multi-object (+7.2\%), dense-crowd (+5.0\%), landscape (+3.3\%), indoor-clutter (+6.9\%), and abstract-art prompts (+13.7\%).
These results indicate that AesMask is not a center-subject shortcut, but follows prompt-conditioned aesthetic affinity even when the foreground/background structure is weak.
The additional runtime diagnostics also support the practical cost profile: on Lumina-Next, full AccelAes reduces latency from 12.37\,s to 5.86\,s (2.11$\times$), while AesMask construction and Spatial-CFG splitting add only 0.18\,s over dense inference.
On the 4-step FLUX.1-schnell distilled DiT, AccelAes improves IR from 1.090 to 1.132 without degrading HPSv2 or CLIP, showing compatibility even when temporal redundancy has already been compressed by distillation.
Finally, dynamic mask refresh does not improve the default static AesMask on hard non-centric prompts; frequent refresh introduces region jitter and lowers both quality and speed.
We also stress tested the main failure cases by mining the worst AccelAes-to-dense ImageReward drops from dense-crowd, indoor-clutter, abstract-art, and landscape prompts.
These cases reveal the expected boundary of early region selection: when attention is diffuse, an overly narrow active region can freeze small details outside the mask.
The late-mask and entropy fallback rules recover 92.7--95.4\% of dense ImageReward without per-prompt tuning.
We therefore keep AesMask static by default and activate corrective expansion only when mask confidence becomes unreliable.
\arxivonly{Additional tables for the human study, prompt robustness, overhead, distilled-model compatibility, and failure mitigation are provided in Appendix~\ref{app:rebuttal_extra}.}

\begin{figure*}[!tbp]
  \centering
  \includegraphics[width=\linewidth]{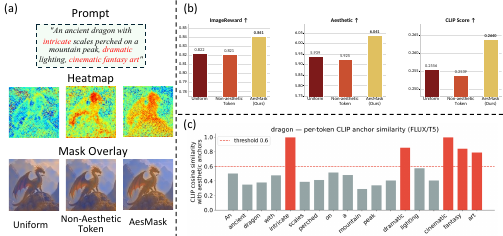}
\caption{\textbf{AesMask analysis and ablation.}
We study how AesMask is constructed and how it affects acceleration and quality. We ablate key design choices such as anchor selection, cross attention aggregation, and mask thresholding, and we report their impact on both runtime and aesthetic metrics.}
  \label{fig:4}
\end{figure*}

\paragraph{Lumina-Next-T2I Comparisons.}
 We provide a strict comparison on the same metric set and protocol in Table~\ref{tab:2}.
At a similar speedup range, several training-free baselines show large drops on ImageReward, which indicates that uniform temporal skipping or generic token reduction can be brittle when preference-oriented quality matters.
For example, TaylorSeer operates in a comparable speed regime but incurs a substantial ImageReward degradation.
$\Delta$-DiT and SDiT further reduce ImageReward despite non-trivial speed gains.
In contrast, AccelAes achieves the best ImageReward while also reaching the largest speedup among listed methods.
This gap suggests that allocating compute and guidance according to aesthetic token affinity is more robust than applying uniform reuse across all regions and steps.

\paragraph{SD3 and FLUX Comparisons.}
Fig.~\ref{fig:3} evaluates the same design on SD3 and FLUX and highlights two practical observations.
First, the relative ranking between methods can change across backbones, which suggests that aggressive temporal reuse is sensitive to the architecture and sampling configuration.
Second, AccelAes remains more stable in preference proxies at practical speedups, which is consistent with its design that preserves updates in aesthetics-relevant regions instead of uniformly replacing Transformer evaluations.
This supports that the benefit of aesthetics-guided spatial allocation is not restricted to Lumina-Next.

\begin{figure*}[!tbp]
  \centering
  \includegraphics[width=\linewidth]{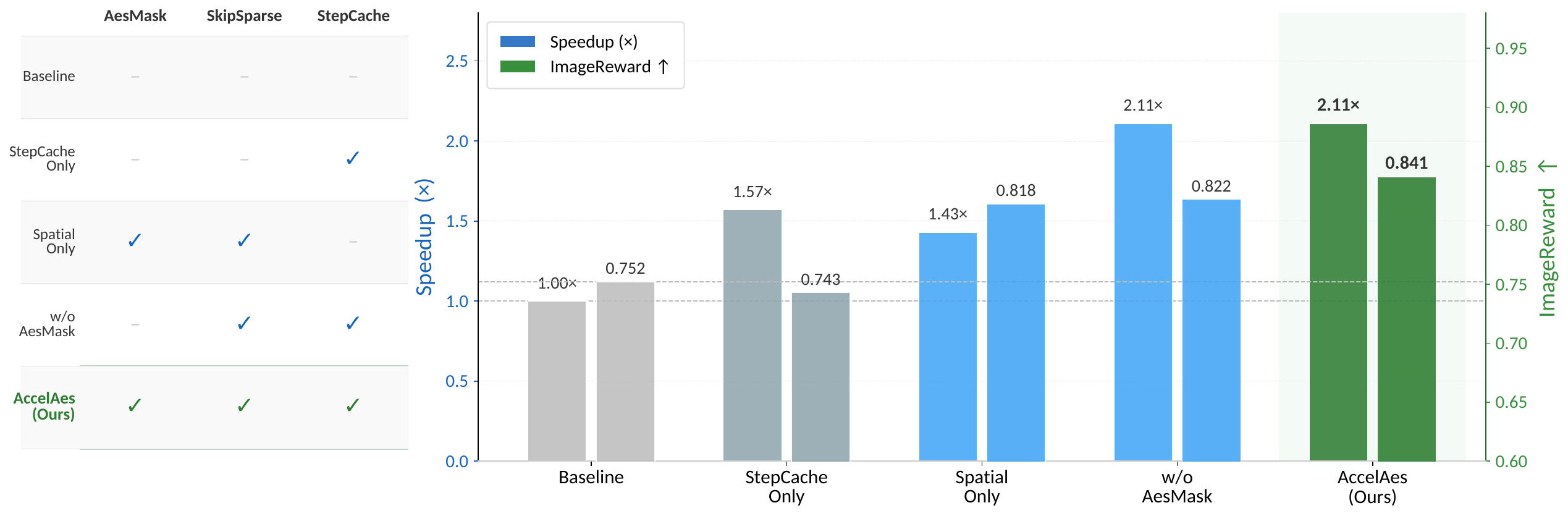}
\caption{\textbf{SkipSparse ablation.}
We evaluate variants that enable spatial allocation, step-level reuse, or both, and report speedup and ImageReward with and without AesMask.}
  \label{fig:5}
\end{figure*}

\subsection{Ablation Results}
\label{sec:ablation}

\paragraph{AesMask Ablation.}
Fig.~\ref{fig:4} compares Uniform masking, a Non-Aesthetic-Token variant, and the proposed AesMask.
In Fig.~\ref{fig:4} (a), Uniform and Non-Aesthetic-Token masks produce less targeted focus regions, while AesMask yields a compact region that aligns with aesthetic descriptors in the prompt.
Fig.~\ref{fig:4} (b) shows that this alignment translates into a more reliable quality profile on ImageReward, Aesthetic score, and CLIP score.
Fig.~\ref{fig:4} (c) further indicates that words with clear aesthetic meaning receive higher importance under anchor matching, which supports our token selection step.
Together, these results suggest that AesMask provides an effective region prior for compute budgeting and region-wise guidance across prompts with different aesthetic emphasis.

\paragraph{SkipSparse Ablation.}
Fig.~\ref{fig:5} shows distinct roles of the two SkipSparse paths.
Step-level reuse achieves a larger speedup (1.57$\times$) but slightly reduces ImageReward (0.743 vs.\ 0.752), which indicates that uniform temporal replacement can be fragile for preference-oriented quality.
Spatial allocation improves ImageReward (0.818) with a moderate speedup (1.43$\times$), suggesting that focusing updates on critical regions is important.
At the same speedup (2.11$\times$), removing AesMask lowers ImageReward (0.822 vs.\ 0.841), showing that AesMask provides a stronger region prior that makes SkipSparse more selective.
Together, these results explain why combining the two paths with AesMask yields the best trade-off.


%

\begin{figure*}[!t]
  \centering
  \includegraphics[width=0.95
\linewidth]{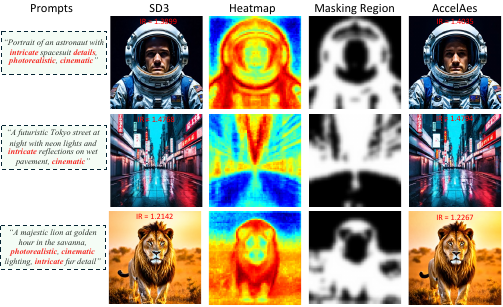}
    \caption{\textbf{Aesthetics-induced spatial non-uniformity and its effect.}
For representative prompts on SD3, we visualize cross-attention heatmaps, the derived aesthetic-focus mask, the selected regions, and the final AccelAes outputs. The localized reallocation of compute and guidance enhances perceptually important details, consistent with improvements in ImageReward (IR) annotated for each example.}
  \label{fig:6}
\end{figure*}
%



\subsection{Visualizations}
\label{sec:visualizations}

\paragraph{Aesthetic Spatial Non-uniformity.}
Fig.~\ref{fig:6} shows cross-attention maps induced by aesthetic descriptors, the resulting AesMask, selected regions, and SD3 outputs. The attention is spatially concentrated on perceptually decisive structures (e.g., faces and high-frequency details) rather than uniform. Thresholding these responses yields a compact mask that suppresses low-impact background regions. AccelAes uses AesMask to reallocate both computation and CFG toward masked tokens, and the annotated IR gains indicate a consistent preference improvement under identical prompts and sampling settings.
\begin{figure*}[!tbp]
  \centering
  \includegraphics[width=0.95
\linewidth]{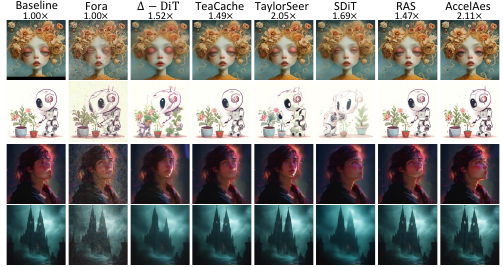}
\caption{\textbf{Qualitative comparison across training-free acceleration methods.}
We compare AccelAes with the dense baseline and representative training-free acceleration baselines under identical prompts. The speedup (relative to the baseline) is shown above each column. AccelAes preserves fine textures and locally salient details more reliably at high speedup, consistent with its superior multi-metric performance in Table~\ref{tab:2}.}
  \label{fig:7}
\end{figure*}
\paragraph{Qualitative Comparison across Acceleration Methods.}
Fig.~\ref{fig:7} compares AccelAes with representative baselines under identical prompts and reports the corresponding speedups. Uniform skipping at higher aggressiveness often blurs fine textures or introduces local artifacts, whereas AccelAes better preserves perceptually decisive details while achieving the largest acceleration in the shown setting. This qualitative evidence is consistent with the quantitative results.
Taken together, the metric comparisons, human preference study, ablations, and visual evidence indicate that AccelAes improves the practical speed--quality frontier across diverse prompt and backbone conditions by protecting aesthetics-relevant regions rather than uniformly reducing computation.
\arxivonly{More qualitative examples and hyperparameter sensitivity visualizations are included in Appendix Figs.~\ref{fig:lumina}--\ref{fig:ratio}.}


%% file: chapters/Conclusion.tex
We presented \textbf{AccelAes}, a training-free acceleration framework for Diffusion Transformers that leverages aesthetics-driven spatial non-uniformity and temporal prediction redundancy in denoising.
By combining \textbf{AesMask}, \textbf{SkipSparse}, and output-level step prediction reuse, AccelAes protects perceptually sensitive tokens while compressing low-impact computation across regions and timesteps.
Across three representative DiT backbones, AccelAes consistently improves the speed--quality trade-off, and ablations, human preferences, and robustness checks confirm the complementarity of spatial adaptivity and temporal reuse.

The broader implication is that training-free DiT acceleration need not be purely temporal caching.
Preserving computation in prompt-relevant regions connects efficiency to aesthetic preference, while the main limitation remains dependence on reliable region estimates when salient regions shift.
This makes AccelAes suitable when preference quality and latency must be optimized together, especially in deployment settings with strict latency budgets.
Future work may study confidence tests and finer-grained refresh schedules.

%% file: chapters/Appendix.tex
\section{Implementation Details}
\label{app:impl}

\subsection{Full Aesthetic Anchor List}
\label{app:anchors}
AccelAes constructs AesMask from prompt semantics and cross-attention signals by using CLIP-based semantic matching~\cite{8}. The implementation uses a fixed vocabulary of \textbf{34 anchor words}, grouped into six semantic categories:
\begin{itemize}
    \item \textbf{Style / rendering quality:} photorealistic, realistic, cinematic, highly detailed, artistic, masterpiece, professional photography, soft bokeh, bokeh, dramatic lighting, fantasy art, studio lighting.
    \item \textbf{Detail / sharpness:} detailed, intricate, sharp focus, sharp.
    \item \textbf{Aesthetic judgment:} stunning, beautiful, elegant, vivid, vibrant.
    \item \textbf{Photography / composition:} depth of field, volumetric lighting, close up, portrait, full body.
    \item \textbf{Subject emphasis:} main subject, foreground, focused.
    \item \textbf{Content-level fallback terms:} subject, character, object, figure, person.
\end{itemize}

Unless otherwise noted, this 34-anchor vocabulary is kept fixed across all experiments.

\subsection{AesMask Hyperparameter Table}
\label{app:aesmask_hparams}
The default AesMask hyperparameters are summarized in Table~\ref{tab:3}. Unless otherwise stated, all experiments use the same configuration, including CLIP ViT-L/14~\cite{8} for token--anchor similarity, the 34-anchor vocabulary from Sec.~\ref{app:anchors}, top-$r$ prompt-token selection by maximum anchor similarity, aggregation over all cross-attention layers, and a default implementation skip ratio of 0.50. The latter corresponds to the 50th percentile ($p=50$) in Eq.~(7) of the main paper. The mask is constructed once at step 5 and reused for the remaining denoising steps.

\subsection{Backbone Adaptation and StepCache Settings}
\label{app:stepcache}

AccelAes is compatible with different DiT-style backbones~\cite{9}, but the exact acceleration path depends on whether the architecture supports spatially localized computation.

For \textbf{Lumina-Next-T2I}~\cite{11}, we use the full AccelAes configuration, including AesMask-guided sparse spatial computation and StepCache-based temporal reuse. This backbone also serves as the main testbed for the SkipSparse ablations in Sec.~\ref{app:skip}.

For \textbf{SD3-Medium}, we adopt a conservative spatial configuration. Since SD3-Medium uses MMDiT-style joint processing of image and text tokens, spatially differentiated CFG~\cite{5} is less stable in our pilot tuning. We therefore use the AesMask-guided spatial path together with temporal StepCache, while keeping CFG spatially uniform; region-wise CFG modulation is disabled.

For \textbf{FLUX.1-dev}~\cite{44}, the final AccelAes configuration mainly uses StepCache-style temporal reuse. Its architecture does not expose the same spatial SkipSparse path as Lumina.

The backbone-specific StepCache settings are summarized in Table~\ref{tab:4}. Across all three backbones, we use a refresh interval of $\Delta=2$, while the total number of denoising steps follows the default setting of each sampler~\cite{3,6}. Under this configuration, the theoretical skip ratio is approximately 46\% for all backbones, while the realized standalone speedup varies across architectures.

\begin{table}[!tbp]
\centering
\small
\setlength{\tabcolsep}{4pt}
\renewcommand{\arraystretch}{1.12}
\caption{Default AesMask hyperparameters.}
\label{tab:3}
\begin{tabular}{p{0.32\linewidth}p{0.24\linewidth}p{0.34\linewidth}}
\toprule
Parameter & Default & Description \\
\midrule
CLIP model &
\begin{tabular}[t]{@{}l@{}}
\texttt{\small clip-vit-large-}\\
\texttt{\small patch14}
\end{tabular}
& token semantic similarity \\
Anchor set size $K$ & 34 & full vocabulary in Sec.~\ref{app:anchors} \\
Aesthetic-token selection & top-$r$ & ranked by maximum anchor similarity \\
Aggregation layer set $L$ & all cross-attn layers & averaged by default \\
\texttt{\small skip\_ratio} $q$ & 0.50 & implementation fraction; $p=100q=50$ in the main-paper notation \\
\texttt{\small mask\_step} & 5 & one-shot mask construction step \\
Region method & threshold & direct percentile thresholding \\
\bottomrule
\end{tabular}
\end{table}

\begin{table}[!tbp]
\centering
\small
\setlength{\tabcolsep}{4pt}
\renewcommand{\arraystretch}{1.12}
\caption{Backbone-specific StepCache settings and standalone speedup statistics.}
\label{tab:4}
\begin{tabular}{lccc}
\toprule
Parameter & Lumina & SD3 & FLUX \\
\midrule
Refresh interval $\Delta$ & 2 & 2 & 2 \\
Warmup steps $T_w$ & 5 & 5 & 5 \\
Total denoising steps $T$ & 30 & 28 & 28 \\
Theoretical skip ratio & 46.7\% & 46.4\% & 46.4\% \\
Standalone StepCache speedup & 1.57$\times$ & 1.50$\times$ & 1.73$\times$ \\
\bottomrule
\end{tabular}
\end{table}

\subsection{Hardware and Evaluation Environment}
\label{app:runtime}

All experiments are conducted on a single NVIDIA GeForce RTX 5090 GPU with batch size 1. Runtime is measured as end-to-end per-image latency, including the full sampling trajectory, classifier-free guidance passes when enabled~\cite{5}, and VAE decoding. Lumina-Next-T2I~\cite{11} and FLUX.1-dev~\cite{44} use BF16 mixed precision, while SD3-Medium uses FP16. The one-time CLIP loading overhead is below 0.05\,s per prompt and is negligible relative to the full generation latency.

\section{Full Quantitative Results}
\label{app:full}
We next report the full quantitative results under the evaluation protocol described in Sec.~\ref{app:runtime}. The compared methods include representative training-free acceleration baselines based on caching, reuse, and forecasting, including $\Delta$-DiT~\cite{14}, FORA~\cite{15}, TeaCache~\cite{17}, and TaylorSeer~\cite{21}.

\subsection{Full Results on SD3-Medium}
\label{app:sd3}

\begin{table}[!tbp]
\centering
\small
\setlength{\tabcolsep}{4pt}
\renewcommand{\arraystretch}{1.10}
\caption{Quantitative comparison on SD3-Medium against representative training-free acceleration baselines~\cite{14,15,17,21}.}
\label{tab:5}
\resizebox{\linewidth}{!}{%
\begin{tabular}{l cc c c c c c}
\toprule
Methods & Time$\downarrow$ & Speedup$\uparrow$ & CLIP$\uparrow$ & IR$\uparrow$ & HPS$\uparrow$ & Aesth$\uparrow$ & Edge$\uparrow$ \\
\midrule
SD3-Medium & 3.52 & 1.00$\times$ & 0.2662 & 0.879 & 0.2895 & 5.733 & 0.916 \\
TeaCache~\cite{17} & 2.49 & 1.41$\times$ & 0.2667 & 0.847 & 0.2863 & 5.717 & 0.881 \\
TaylorSeer~\cite{21} & 1.74 & 2.02$\times$ & 0.2624 & 0.729 & 0.2723 & 5.561 & 0.998 \\
$\Delta$-DiT~\cite{14} & 2.33 & 1.51$\times$ & 0.2634 & 0.828 & 0.2804 & 5.656 & 0.850 \\
FORA~\cite{15} & 3.54 & 0.99$\times$ & 0.2663 & 0.703 & 0.2643 & 5.522 & 0.830 \\
\oursrow AccelAes (ours) & 2.34 & 1.50$\times$ & 0.2744 & 0.904 & 0.3014 & 5.990 & 0.944 \\
\bottomrule
\end{tabular}
}
\end{table}

Table~\ref{tab:5} reports the full comparison on SD3-Medium against TeaCache~\cite{17}, TaylorSeer~\cite{21}, $\Delta$-DiT~\cite{14}, and FORA~\cite{15}. AccelAes achieves a practical 1.50$\times$ speedup while obtaining the best preference-oriented metrics among the accelerated methods, including the highest ImageReward~\cite{34} (0.904), the highest HPS~\cite{41} (0.3014), and the highest aesthetic score (5.990). More aggressive temporal extrapolation, such as TaylorSeer~\cite{21}, achieves higher nominal speedup but causes a clear degradation in preference quality. These results suggest that preserving updates in aesthetically salient regions is more reliable than uniformly replacing denoising steps on SD3-Medium.

\subsection{Full Results on FLUX.1-dev}
\label{app:flux}

\begin{table}[!tbp]
\centering
\small
\setlength{\tabcolsep}{4pt}
\renewcommand{\arraystretch}{1.10}
\caption{Quantitative comparison on FLUX.1-dev against representative training-free acceleration baselines~\cite{17,21}.}
\label{tab:6}
\resizebox{\linewidth}{!}{%
\begin{tabular}{l cc c c c c c}
\toprule
Methods & Time$\downarrow$ & Speedup$\uparrow$ & CLIP$\uparrow$ & IR$\uparrow$ & HPS$\uparrow$ & Aesth$\uparrow$ & Edge$\uparrow$ \\
\midrule
FLUX.1-dev~\cite{44} & 12.66 & 1.00$\times$ & 0.2753 & 1.233 & 0.3200 & 6.243 & 0.560 \\
TeaCache~\cite{17} & 5.92 & 2.14$\times$ & 0.2777 & 1.225 & 0.3154 & 6.301 & 0.529 \\
TaylorSeer~\cite{21} & 5.96 & 2.12$\times$ & 0.2765 & 1.304 & 0.3217 & 6.309 & 0.572 \\
\oursrow AccelAes (ours) & 7.31 & 1.73$\times$ & 0.2772 & 1.317 & 0.3214 & 6.372 & 0.590 \\
\bottomrule
\end{tabular}
}
\end{table}

Table~\ref{tab:6} reports the quantitative comparison on FLUX.1-dev~\cite{44}. Since FLUX.1-dev does not support our spatially localized SkipSparse path in the same form as Lumina-Next-T2I~\cite{11}, the final AccelAes configuration on this backbone mainly reflects the benefit of StepCache-style temporal reuse. Under this setting, AccelAes reaches 1.73$\times$ speedup and achieves the best ImageReward~\cite{34} (1.317), the best aesthetic score (6.372), and the highest Edge score (0.590) among the compared methods. TeaCache~\cite{17} and TaylorSeer~\cite{21} are faster on this backbone, but their overall quality is less balanced across metrics.

\section{Supplementary Ablation Studies}

\subsection{Full SkipSparse Ablation}
\label{app:skip}
\begin{table}[!tbp]
\centering
\small
\setlength{\tabcolsep}{4pt}
\renewcommand{\arraystretch}{1.10}
\caption{SkipSparse ablation on Lumina-Next-T2I~\cite{11}.}
\label{tab:7}
\begin{tabular}{lccccc}
\toprule
Setting & Speedup$\uparrow$ & CLIP$\uparrow$ & IR$\uparrow$ & HPS$\uparrow$ & Aesth$\uparrow$ \\
\midrule
Baseline & 1.00$\times$ & 0.2531 & 0.752 & 0.2710 & 5.941 \\
StepCache Only & 1.57$\times$ & 0.2500 & 0.743 & 0.2723 & 5.949 \\
Spatial Only (attn) & 1.43$\times$ & 0.2545 & 0.818 & 0.2725 & 5.920 \\
Spatial + FFN & 1.43$\times$ & 0.2590 & 0.825 & 0.2734 & 5.957 \\
\oursrow \textbf{AccelAes (Full)} & 2.11$\times$ & 0.2640 & 0.841 & 0.2740 & 6.041 \\
\bottomrule
\end{tabular}
\end{table}

Table~\ref{tab:7} presents the full SkipSparse ablation on Lumina-Next-T2I~\cite{11}. StepCache alone contributes the larger single-module speedup, improving latency from 1.00$\times$ to 1.57$\times$, but it slightly reduces ImageReward~\cite{34} relative to the dense baseline (0.743 vs.\ 0.752). In contrast, the spatial path alone provides a smaller speedup (1.43$\times$) but substantially improves ImageReward to 0.818. Adding FFN sparsification on top of spatial attention yields only modest additional gains over the attention-only spatial variant in the current setting, which suggests that the main spatial benefit already comes from selectively updating attention-dominant regions. Combining the spatial path with StepCache yields the strongest overall result: AccelAes reaches 2.11$\times$ speedup while also achieving the best CLIP~\cite{8}, ImageReward~\cite{34}, HPS~\cite{41}, and Aesthetic scores in Table~\ref{tab:7}.

\subsection{Full Comparison of AesMask Token Weighting Strategies}

\begin{table}[!tbp]
\centering
\small
\setlength{\tabcolsep}{4pt}
\renewcommand{\arraystretch}{1.10}
\caption{Comparison of token weighting strategies for AesMask construction.}
\label{tab:8}
\begin{tabular}{lccc}
\toprule
Token weighting strategy & IR$\uparrow$ & Aesthetic$\uparrow$ & CLIP$\uparrow$ \\
\midrule
Uniform & 0.8222 & 5.9386 & 0.2556 \\
Non-Aesthetic & 0.8207 & 5.9252 & 0.2539 \\
\oursrow \textbf{AesMask (Ours)} & 0.8410 & 6.0414 & 0.2640 \\
\bottomrule
\end{tabular}
\end{table}

Table~\ref{tab:8} compares three token-weighting strategies for AesMask construction: uniform weighting, non-aesthetic-token weighting, and the proposed anchor-aware AesMask weighting based on CLIP~\cite{8}. The proposed strategy achieves the best performance on all three reported metrics, improving ImageReward~\cite{34} to 0.8410, Aesthetic score to 6.0414, and CLIP score to 0.2640. By comparison, uniform weighting and non-aesthetic-token weighting produce weaker preference gains and lower alignment scores. These results support the use of semantically matched aesthetic anchors as a more effective prior for region selection than treating all prompt tokens equally or suppressing aesthetic descriptors.

\subsection{Sensitivity to \texttt{skip\_ratio} $q$}
The implementation parameter $q\in(0,1)$ is related to the percentile notation in the main paper by $p=100q$. Thus, the default \texttt{skip\_ratio}$=0.50$ corresponds to $p=50$.
\begin{table}[!tbp]
\centering
\small
\setlength{\tabcolsep}{4pt}
\renewcommand{\arraystretch}{1.10}
\caption{Sensitivity to \texttt{skip\_ratio}.}
\label{tab:9}
\begin{tabular}{lcccc}
\toprule
\texttt{skip\_ratio} & Speedup$\uparrow$ & IR$\uparrow$ & CLIP$\uparrow$ & HPS$\uparrow$ \\
\midrule
0.30 & $\sim$2.11$\times$ & 0.8415 & 0.2595 & 0.2738 \\
0.40 & $\sim$2.11$\times$ & 0.8334 & 0.2542 & 0.2731 \\
\oursrow 0.50 & $\sim$2.11$\times$ & 0.8410 & 0.2640 & 0.2740 \\
0.60 & $\sim$2.11$\times$ & 0.7990 & 0.2541 & 0.2724 \\
0.70 & $\sim$2.11$\times$ & 0.7643 & 0.2532 & 0.2715 \\
\bottomrule
\end{tabular}
\end{table}

The sensitivity to the spatial skip ratio is summarized quantitatively in Table~\ref{tab:9} and illustrated qualitatively in Fig.~\ref{fig:ratio}. The overall speedup remains nearly unchanged across the tested settings because the dominant latency reduction still comes from the temporal reuse path, consistent with the caching-based acceleration literature~\cite{13,14,15,17,21}. However, generation quality is clearly sensitive to the choice of skip ratio. A smaller ratio such as 0.30 yields the highest ImageReward~\cite{34} (0.8415), but 0.50 provides a better overall balance across ImageReward, CLIP~\cite{8}, and HPS~\cite{41}. As the skip ratio increases beyond 0.50, all quality metrics degrade steadily. The visual examples in Fig.~\ref{fig:ratio} show the same pattern: moderate sparsity preserves detail and structure more reliably, whereas overly aggressive sparsity gradually weakens local fidelity. We therefore use 0.50 as the default setting.

\subsection{Sensitivity to \texttt{mask\_step}}
\begin{table}[!tbp]
\centering
\small
\setlength{\tabcolsep}{4pt}
\renewcommand{\arraystretch}{1.10}
\caption{Sensitivity to \texttt{mask\_step}.}
\label{tab:10}
\begin{tabular}{lcccc}
\toprule
\texttt{mask\_step} & Speedup$\uparrow$ & IR$\uparrow$ & CLIP$\uparrow$ & HPS$\uparrow$ \\
\midrule
3  & 2.27$\times$ & 0.832 & 0.2633 & 0.2740 \\
\oursrow 5  & 2.11$\times$ & 0.841 & 0.2640 & 0.2740 \\
7  & 1.92$\times$ & 0.818 & 0.2637 & 0.2733 \\
10 & 1.73$\times$ & 0.818 & 0.2615 & 0.2728 \\
15 & 1.42$\times$ & 0.783 & 0.2604 & 0.2718 \\
\bottomrule
\end{tabular}
\end{table}

Table~\ref{tab:10} and Fig.~\ref{fig:maskstep} study the effect of the one-shot mask construction step. Earlier masking leads to stronger acceleration, with mask\_step=3 reaching 2.27$\times$ speedup, but the best ImageReward~\cite{34} is obtained at mask\_step=5 (0.841). As the mask is delayed further, both speedup and preference-oriented quality gradually decline. The qualitative examples in Fig.~\ref{fig:maskstep} are consistent with the quantitative trend in Table~\ref{tab:10}: constructing the mask too early or too late can reduce local detail quality, whereas mask\_step=5 provides the most stable visual trade-off. Based on this evidence, we adopt mask\_step=5 as the default setting throughout the main experiments.

\subsection{Why Spatial CFG Is Disabled on SD3-Medium}

For SD3-Medium, we disable spatially differentiated CFG~\cite{5} in the final configuration. In our pilot tuning, this design did not provide a consistent improvement at matched latency and was less stable than the uniform CFG setting. A likely reason is that SD3-Medium uses MMDiT-style joint processing of image and text tokens, where text conditioning is already deeply integrated into the denoising blocks. Under this architecture, additional region-wise CFG scaling can introduce unnecessary scale inconsistency across spatial locations, a phenomenon related to recent observations on spatial inconsistency in guidance~\cite{35}. We therefore retain the simpler uniform guidance configuration for all reported SD3-Medium results.

\section{Runtime Decomposition}
\begin{table}[!tbp]
\centering
\small
\setlength{\tabcolsep}{4pt}
\renewcommand{\arraystretch}{1.10}
\caption{Runtime decomposition on Lumina-Next-T2I~\cite{11}.}
\label{tab:11}
\begin{tabular}{lccc}
\toprule
Component & Time$\downarrow$ & Speedup$\uparrow$ & Relative saving \\
\midrule
Baseline & 12.37s & 1.00$\times$ & --- \\
StepCache only & 7.87s & 1.57$\times$ & 36.4\% \\
Spatial sparse only & 8.65s & 1.43$\times$ & 30.1\% \\
\oursrow AccelAes (full) & 5.86s & 2.11$\times$ & 52.6\% \\
CLIP one-time overhead & $<$0.05s & negligible & --- \\
\bottomrule
\end{tabular}
\end{table}

Table~\ref{tab:11} decomposes the runtime contribution of each component on Lumina-Next-T2I~\cite{11}. StepCache alone reduces latency from 12.37\,s to 7.87\,s, corresponding to a 36.4\% relative saving, while the spatial sparse path alone reduces latency to 8.65\,s, corresponding to a 30.1\% saving. Combining both components yields the full AccelAes latency of 5.86\,s and a total speedup of 2.11$\times$, which corresponds to a 52.6\% reduction relative to the dense baseline. The gain is smaller than the arithmetic sum of the individual savings because the two mechanisms partially overlap on the same denoising steps. Nevertheless, Table~\ref{tab:11} shows that the combined design remains clearly stronger than either component alone. This trend is consistent with prior findings that different training-free acceleration mechanisms can be complementary rather than redundant~\cite{13,14,17,21}.

\section{Rebuttal-Driven Additional Experiments}
\label{app:rebuttal_extra}

This section consolidates additional experiments conducted during rebuttal into the camera-ready supplementary material. These experiments target three reviewer concerns: whether the gains are human-perceived rather than metric-specific, whether the method generalizes beyond center-subject prompts, and whether the default static AesMask is a reasonable design choice.

\subsection{Blind Human Preference Study}

\begin{table}[!tbp]
\centering
\small
\setlength{\tabcolsep}{4pt}
\renewcommand{\arraystretch}{1.10}
\caption{Blind pairwise human preference study. Win rates exclude ties. Each annotator answered four questions for each pair: overall preference, prompt alignment, visual appeal, and fewer artifacts.}
\label{tab:supp_human_pref}
\resizebox{\linewidth}{!}{%
\begin{tabular}{lcccc}
\toprule
Comparison & Overall & Alignment & Appeal & Fewer artifacts \\
\midrule
\oursrow AccelAes vs.\ Dense & 67\% & 59\% & 68\% & 64\% \\
\oursrow AccelAes vs.\ FORA & \goodcell{89\%} & \goodcell{90\%} & \goodcell{87\%} & \goodcell{94\%} \\
\oursrow AccelAes vs.\ w/o Spatial-CFG & 66\% & 69\% & 66\% & 68\% \\
\bottomrule
\end{tabular}
}
\end{table}

We collected 15 annotators $\times$ 60 pairs $\times$ 4 questions with randomized left/right order. AccelAes is preferred over the dense baseline, FORA, and the variant without Spatial-CFG across the four questions. Binomial tests on non-tie votes reject the 50\% chance null for the main preference comparisons ($p<0.001$, except alignment against the dense baseline), indicating that the gain is not solely an artifact of ImageReward or HPSv2.

\subsection{Prompt and Scene Robustness}

\begin{table}[!tbp]
\centering
\small
\setlength{\tabcolsep}{4pt}
\renewcommand{\arraystretch}{1.10}
\caption{Prompt-wording robustness on 200 base prompts. AccelAes improves ImageReward with and without explicit aesthetic descriptors.}
\label{tab:supp_prompt_robust}
\begin{tabular}{lccc}
\toprule
Prompt setting & Dense IR & AccelAes IR & Relative change \\
\midrule
Descriptor-free & 1.109 & \goodcell{1.144} & +3.1\% \\
Descriptor-added & 1.168 & \goodcell{1.224} & +4.8\% \\
\bottomrule
\end{tabular}
\end{table}

\begin{table}[!tbp]
\centering
\small
\setlength{\tabcolsep}{4pt}
\renewcommand{\arraystretch}{1.10}
\caption{Complex-scene robustness. Each category uses hard prompts with two seeds per prompt. All categories show positive ImageReward gains.}
\label{tab:supp_scene_robust}
\begin{tabular}{lcc@{\quad}lcc}
\toprule
Category & Dense IR & $\Delta$IR & Category & Dense IR & $\Delta$IR \\
\midrule
single-subj. & 1.333 & +1.8\% & landscape & 0.836 & +3.3\% \\
multi-object & 0.791 & +7.2\% & indoor-clutter & 0.518 & +6.9\% \\
dense-crowd & 0.436 & +5.0\% & abstract-art & 0.233 & \goodcell{+13.7\%} \\
\bottomrule
\end{tabular}
\end{table}

The descriptor-free setting confirms that AccelAes does not require prompt engineering with explicit aesthetic adjectives. The complex-scene study further shows that improvements persist for dense crowds, landscapes, indoor clutter, and abstract art. The largest gain appears on abstract art, where foreground/background structure is least explicit. This indicates that AesMask follows prompt-conditioned aesthetic affinity rather than a fixed center-subject prior.

\subsection{End-to-End Overhead and Distilled-Model Compatibility}

\begin{table}[!tbp]
\centering
\small
\setlength{\tabcolsep}{4pt}
\renewcommand{\arraystretch}{1.10}
\caption{End-to-end overhead diagnostic on Lumina-Next-T2I. All scheduling, masking, Spatial-CFG splitting, and decoding costs are included.}
\label{tab:supp_overhead}
\begin{tabular}{lcc}
\toprule
Configuration & Latency (s) & Speedup \\
\midrule
Dense baseline & 12.37 & 1.00$\times$ \\
StepCache only & 7.87 & 1.57$\times$ \\
SkipSparse spatial path only & 8.65 & 1.43$\times$ \\
AesMask + Spatial-CFG overhead (dense compute) & 12.55 & 0.99$\times$ \\
\oursrow Full AccelAes & \goodcell{5.86} & \goodcell{2.11$\times$} \\
\bottomrule
\end{tabular}
\end{table}

\begin{table}[!tbp]
\centering
\small
\setlength{\tabcolsep}{4pt}
\renewcommand{\arraystretch}{1.10}
\caption{Compatibility check on FLUX.1-schnell, a 4-step distilled DiT. The test focuses on quality compatibility because distillation already compresses temporal redundancy.}
\label{tab:supp_schnell}
\begin{tabular}{lccc}
\toprule
Configuration & IR & HPSv2 & CLIP \\
\midrule
schnell-dense & 1.090 & 0.301 & 0.268 \\
schnell + TaylorSeer & 1.100 & 0.301 & 0.269 \\
\oursrow schnell + AccelAes & \goodcell{1.132} & \goodcell{0.304} & \goodcell{0.270} \\
\bottomrule
\end{tabular}
\end{table}

The overhead diagnostic shows that AesMask construction and Spatial-CFG splitting add only 0.18\,s, or 1.4\% of dense latency, while the full system reduces latency from 12.37\,s to 5.86\,s (2.11$\times$). On FLUX.1-schnell, AccelAes improves ImageReward by 3.9\% without degrading HPSv2 or CLIP, showing compatibility with a distilled DiT where temporal redundancy is already compressed.

\subsection{Static AesMask and Failure Mitigation}

\begin{table}[!tbp]
\centering
\small
\setlength{\tabcolsep}{4pt}
\renewcommand{\arraystretch}{1.10}
\caption{Static versus dynamic AesMask refresh on hard non-centric prompts. The default static mask is faster and obtains higher ImageReward than dynamic refresh variants.}
\label{tab:supp_static_mask}
\begin{tabular}{lcc}
\toprule
Mask update strategy & Speedup & IR \\
\midrule
\oursrow Static AesMask (default) & \goodcell{2.08$\times$} & \goodcell{0.407} \\
5-step refresh & 1.94$\times$ & 0.392 \\
Per-step refresh & 1.82$\times$ & 0.381 \\
\bottomrule
\end{tabular}
\end{table}

Frequent mask refresh is not automatically more adaptive. Late-stage attention can become noisier and introduce region jitter, which hurts both quality and speed. We therefore keep a static one-shot AesMask by default. For rare failure cases, we mined the worst AccelAes-to-dense ImageReward drops from dense-crowd, indoor-clutter, abstract-art, and landscape prompts. Fixed delayed-mask and fallback variants recover 92.7--95.4\% of dense ImageReward without tuning, indicating that the observed failures are recoverable through active-region correction.

\section{Anchor Coverage Analysis}
For this analysis only, we define a diagnostic anchor hit when a prompt token has maximum anchor similarity of at least 0.60. This threshold is used solely to quantify vocabulary coverage; the inference-time AesMask procedure follows the top-$r$ token selection rule described in the main paper and does not use 0.60 as a gating threshold. We then analyze how often such diagnostic hits occur and which anchors are matched most frequently on prompts drawn from Pick-a-Pic~\cite{33}.

\subsection{Diagnostic Anchor-Hit Statistics}

Table~\ref{tab:12} shows that diagnostic anchor hits occur for 44.9\% of the full Pick-a-Pic v1 unique prompt pool (17,303/38,522) and 65.48\% of the 10,000-prompt evaluation subset (6,548/10,000). This indicates that the anchor vocabulary has broad coverage and is more frequently matched on the aesthetics-oriented subset used in our experiments.

\begin{table}[!tbp]
\centering
\small
\setlength{\tabcolsep}{4pt}
\renewcommand{\arraystretch}{1.10}
\caption{Diagnostic anchor-hit statistics across prompt sets sampled from Pick-a-Pic~\cite{33}. The 0.60 threshold is used only for this coverage analysis.}
\label{tab:12}
\resizebox{\linewidth}{!}{%
\begin{tabular}{lcccc}
\toprule
Set & Total prompts & Hit prompts & Hit rate & Avg. matched anchors \\
\midrule
Pick-a-Pic v1 (unique)~\cite{33} & 38,522 & 17,303 & 44.9\% & 3.47 \\
Main 10k evaluation prompts & 10,000 & 6,548 & 65.48\% & 6.89 \\
\bottomrule
\end{tabular}
}
\end{table}

\subsection{Most Frequent Anchor Hits}

The most frequent matched anchors on the Pick-a-Pic unique prompt pool~\cite{33} are listed in Table~\ref{tab:13}. The dominant terms, including \emph{photorealistic}, \emph{detailed}, \emph{artistic}, \emph{highly detailed}, and \emph{realistic}, are all closely related to rendering quality or perceptual style. This supports the design of the anchor vocabulary and helps explain why anchor-aware weighting improves AesMask quality in practice.

\begin{table}[!tbp]
\centering
\small
\setlength{\tabcolsep}{4pt}
\renewcommand{\arraystretch}{1.10}
\caption{Most frequent anchors on the Pick-a-Pic unique prompt pool~\cite{33}.}
\label{tab:13}
\begin{tabular}{lcc}
\toprule
Anchor & Count & Ratio within triggered prompts \\
\midrule
photorealistic & 7,108 & 41.1\% \\
detailed & 4,939 & 28.5\% \\
artistic & 4,318 & 25.0\% \\
highly detailed & 4,196 & 24.3\% \\
realistic & 3,261 & 18.8\% \\
beautiful & 2,746 & 15.9\% \\
stunning & 2,721 & 15.7\% \\
cinematic & 2,316 & 13.4\% \\
portrait & 1,982 & 11.5\% \\
fantasy art & 1,625 & 9.4\% \\
\bottomrule
\end{tabular}
\end{table}

\subsection{Prompts without Diagnostic Anchor Hits}

A prompt with no anchor hit under the diagnostic 0.60 threshold is not assigned a separate inference-time fallback mode. AesMask still ranks prompt tokens by their anchor-similarity scores and selects the top-$r$ tokens, exactly as specified in the main paper. The 0.60 threshold is used only for the coverage statistics above and does not change the generation procedure.

\section{Additional Qualitative Results}

\begin{figure}[!tbp]
  \centering
  \includegraphics[width=0.95\linewidth]{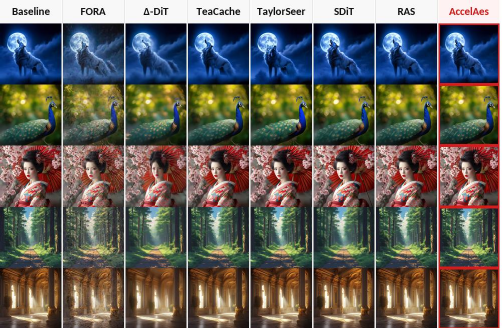}
\caption{\textbf{Additional qualitative comparison on Lumina-Next-T2I~\cite{11}.}
We compare AccelAes with the dense baseline and representative training-free acceleration baselines~\cite{14,15,17,21} under identical prompts and sampling settings. AccelAes more reliably preserves fine textures, subject boundaries, and local lighting structure.}
  \label{fig:lumina}
\end{figure}

\begin{figure}[!tbp]
  \centering
  \includegraphics[width=0.95\linewidth]{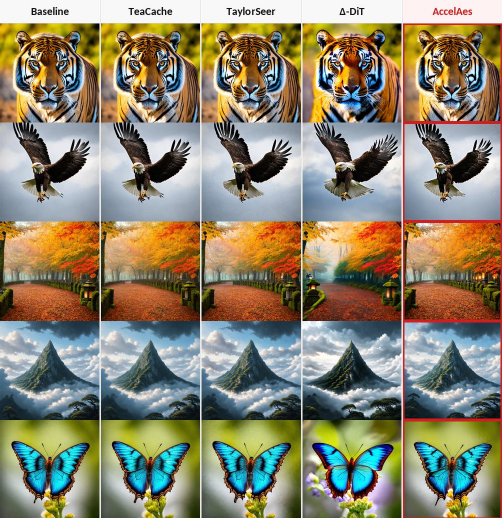}
\caption{\textbf{Additional qualitative comparison on SD3-Medium.}
Under matched prompts and sampling settings, AccelAes better maintains fine-grained structures, local texture fidelity, and perceptually important details than competing baselines~\cite{14,15,17,21}.}
  \label{fig:sd3}
\end{figure}

\begin{figure}[!tbp]
  \centering
  \includegraphics[width=0.95\linewidth]{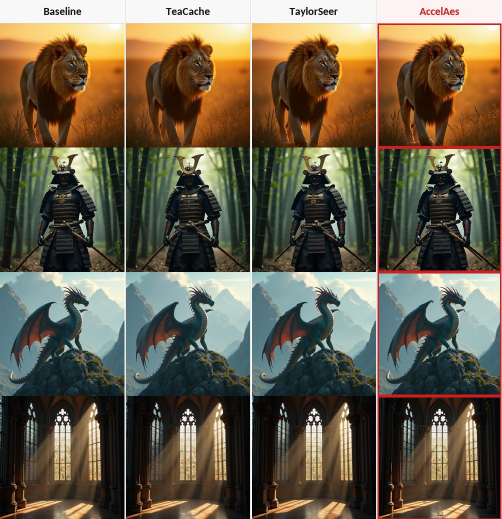}
\caption{\textbf{Additional qualitative comparison on FLUX.1-dev~\cite{44}.}
Although the FLUX.1-dev configuration mainly uses the temporal reuse path, AccelAes still preserves perceptually important details more consistently than competing baselines~\cite{17,21} at practical speedups.}
  \label{fig:flux}
\end{figure}

This section provides additional qualitative evidence complementary to the quantitative results in the main paper. Figs.~\ref{fig:lumina},~\ref{fig:sd3},~\ref{fig:flux} present extra comparisons on Lumina-Next-T2I~\cite{11}, SD3-Medium, and FLUX.1-dev~\cite{44}, respectively. Across these backbones, AccelAes more reliably preserves subject boundaries, local textures, lighting structure, and other perceptually important details than competing baselines under matched prompts and sampling settings.

\begin{figure}[!tbp]
  \centering
  \includegraphics[width=0.95\linewidth]{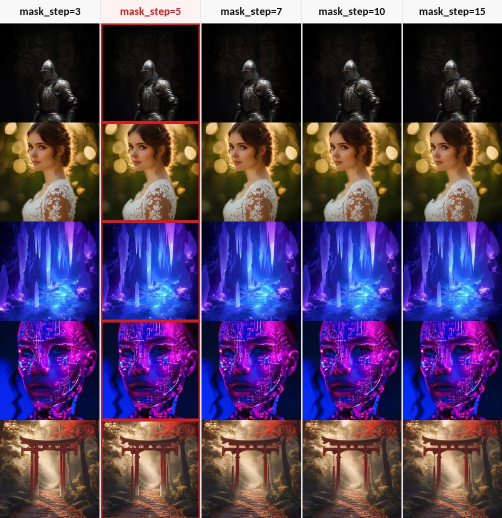}
\caption{\textbf{Qualitative sensitivity to \texttt{mask\_step}.}
We compare different one-shot mask construction steps on representative prompts. Constructing the mask too early or too late weakens the balance between acceleration and visual fidelity, while the default setting \texttt{mask\_step}=5 yields the most stable visual quality across diverse scenes.}
  \label{fig:maskstep}
\end{figure}

\begin{figure}[!tbp]
  \centering
  \includegraphics[width=0.95\linewidth]{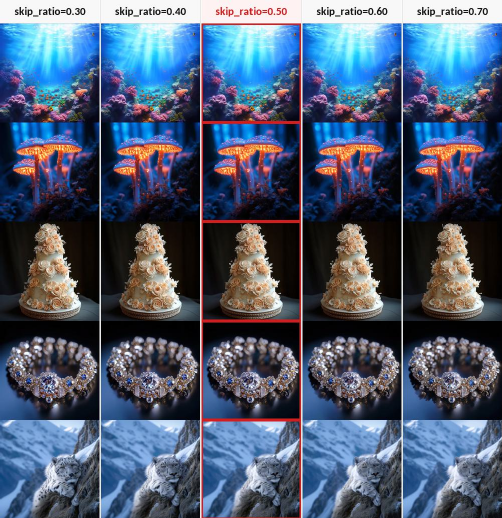}
\caption{\textbf{Qualitative sensitivity to \texttt{skip\_ratio}.}
We compare different spatial skip ratios on representative prompts. Moderate sparsity preserves salient structures, local textures, and perceptually important details more reliably, whereas overly aggressive token skipping gradually degrades visual fidelity. The default setting \texttt{skip\_ratio}=0.50 provides the best overall trade-off.}
  \label{fig:ratio}
\end{figure}

Figs.~\ref{fig:maskstep} and~\ref{fig:ratio} illustrate the sensitivity to two key hyperparameters. Fig.~\ref{fig:maskstep} shows that very early or delayed mask construction weakens visual fidelity, while \texttt{mask\_step}=5 provides the most stable trade-off. Fig.~\ref{fig:ratio} shows that moderate sparsity preserves salient structures and local details more reliably than overly aggressive token skipping. These qualitative observations are consistent with the quantitative trends in Tables~\ref{tab:9} and~\ref{tab:10}.

%% file: main.bib
@inproceedings{1,
  title     = {Denoising Diffusion Probabilistic Models},
  author    = {Ho, Jonathan and Jain, Ajay and Abbeel, Pieter},
  booktitle = {Advances in Neural Information Processing Systems (NeurIPS)},
  year      = {2020},
  pages = {6840--6851}
}

@inproceedings{2,
  title     = {Improved Denoising Diffusion Probabilistic Models},
  author    = {Nichol, Alexander Quinn and Dhariwal, Prafulla},
  booktitle = {Proceedings of the 38th International Conference on Machine Learning (ICML)},
  series    = {Proceedings of Machine Learning Research},
  volume    = {139},
  pages     = {8162--8171},
  editor    = {Meila, Marina and Zhang, Tong},
  year      = {2021},
  publisher = {PMLR}
}

@inproceedings{3,
  title     = {Denoising Diffusion Implicit Models},
  author    = {Song, Jiaming and Meng, Chenlin and Ermon, Stefano},
  booktitle = {International Conference on Learning Representations (ICLR)},
  year      = {2021}
}

@inproceedings{4,
  title     = {High-Resolution Image Synthesis with Latent Diffusion Models},
  author    = {Rombach, Robin and Blattmann, Andreas and Lorenz, Dominik and Esser, Patrick and Ommer, Bjorn},
  booktitle = {Proceedings of the IEEE/CVF Conference on Computer Vision and Pattern Recognition (CVPR)},
  year      = {2022},
  pages = {10684--10695}
}

@article{5,
  title   = {Classifier-Free Diffusion Guidance},
  author  = {Ho, Jonathan and Salimans, Tim},
  journal = {arXiv preprint arXiv:2207.12598},
  year    = {2022}
}

@inproceedings{6,
  title     = {DPM-Solver: A Fast ODE Solver for Diffusion Probabilistic Model Sampling in Around 10 Steps},
  author    = {Lu, Cheng and Zhou, Yuhao and Bao, Fan and Chen, Jianfei and Li, Chongxuan and Zhu, Jun},
  booktitle = {Advances in Neural Information Processing Systems (NeurIPS)},
  year      = {2022},
  pages = {5775--5787}
}

@article{ctrlzsampling,
  title   = {Ctrl-Z Sampling: Scaling Diffusion Sampling with Controlled Random Zigzag Explorations},
  author  = {Mao, Shunqi and Guo, Wei and Zhang, Chaoyi and Long, Jieting and Xie, Ke and Cai, Weidong},
  journal = {arXiv preprint arXiv:2506.20294},
  year    = {2025}
}

@inproceedings{7,
  title     = {FlashAttention: Fast and Memory-Efficient Exact Attention with IO-Awareness},
  author    = {Dao, Tri and Fu, Dan and Ermon, Stefano and Rudra, Atri and R{\'e}, Christopher},
  booktitle = {Advances in Neural Information Processing Systems (NeurIPS)},
  year      = {2022},
pages = {16344--16359}
}

@inproceedings{8,
  title     = {Learning Transferable Visual Models From Natural Language Supervision},
  author    = {Radford, Alec and Kim, Jong Wook and Hallacy, Chris and Ramesh, Aditya and Goh, Gabriel and Agarwal, Sandhini and Sastry, Girish and Askell, Amanda and Mishkin, Pamela and Clark, Jack and Krueger, Gretchen and Sutskever, Ilya},
  booktitle = {Proceedings of the 38th International Conference on Machine Learning (ICML)},
  series    = {Proceedings of Machine Learning Research},
  volume    = {139},
  pages     = {8748--8763},
  editor    = {Meila, Marina and Zhang, Tong},
  year      = {2021},
  publisher = {PMLR}
}

@inproceedings{9,
  title     = {Scalable Diffusion Models with Transformers},
  author    = {Peebles, William and Xie, Saining},
  booktitle = {Proceedings of the IEEE/CVF International Conference on Computer Vision (ICCV)},
  year      = {2023},
  pages = {4195--4205}
}

@inproceedings{10,
  title = {PixArt\mbox{-}$\alpha$: Fast Training of Diffusion Transformer for Photorealistic Text-to-Image Synthesis},
  author    = {Chen, Junsong and Yu, Jincheng and Ge, Chongjian and Yao, Lewei and Xie, Enze and Wu, Yue and Wang, Zhongdao and Kwok, James and Luo, Ping and Lu, Huchuan and Li, Zhenguo},
  booktitle = {International Conference on Learning Representations (ICLR)},
  year      = {2024}
}

@inproceedings{11,
  title     = {Lumina-Next: Making Lumina-T2X Stronger and Faster with Next-DiT},
  author    = {Zhuo, Le and Du, Ruoyi and Xiao, Han and Li, Yangguang and Liu, Dongyang and Huang, Rongjie and Liu, Wenze and Zhu, Xiangyang and Wang, Fu-Yun and Ma, Zhanyu and Luo, Xu and Wang, Zehan and Zhang, Kaipeng and Zhao, Lirui and Liu, Si and Yue, Xiangyu and Ouyang, Wanli and Qiao, Yu and Li, Hongsheng and Gao, Peng},
  booktitle = {Advances in Neural Information Processing Systems (NeurIPS)},
  year      = {2024}
}

@inproceedings{12,
  title     = {Lumina-Image 2.0: A Unified and Efficient Image Generative Framework},
  author    = {Qin, Qi and Zhuo, Le and Xin, Yi and Du, Ruoyi and Li, Zhen and Fu, Bin and Lu, Yiting and Li, Xinyue and Liu, Dongyang and Zhu, Xiangyang and others},
  booktitle = {Proceedings of the IEEE/CVF International Conference on Computer Vision (ICCV)},
  year      = {2025},
  pages     = {20031--20042}
}

@inproceedings{13,
  title     = {DeepCache: Accelerating Diffusion Models for Free},
  author    = {Ma, Xinyin and Fang, Gongfan and Wang, Xinchao},
  booktitle = {Proceedings of the IEEE/CVF Conference on Computer Vision and Pattern Recognition (CVPR)},
  year      = {2024},
  pages = {15762--15772} 
}

@article{14,
  title   = {$\Delta$-DiT: A Training-Free Acceleration Method Tailored for Diffusion Transformers},
  author  = {Chen, Pengtao and Shen, Mingzhu and Ye, Peng and Cao, Jianjian and Tu, Chongjun and Bouganis, Christos-Savvas and Zhao, Yiren and Chen, Tao},
  journal = {arXiv preprint arXiv:2406.01125},
  year    = {2024}
}

@article{15,
  title   = {FORA: Fast-Forward Caching in Diffusion Transformer Acceleration},
  author  = {Selvaraju, Pratheba and Ding, Tianyu and Chen, Tianyi and Zharkov, Ilya and Liang, Luming},
  journal = {arXiv preprint arXiv:2407.01425},
  year    = {2024}
}

@inproceedings{16,
  title     = {Accelerating Diffusion Transformer via Increment-Calibrated Caching with Channel-Aware Singular Value Decomposition},
  author    = {Chen, Zhiyuan and Li, Keyi and Jia, Yifan and Ye, Le and Ma, Yufei},
  booktitle = {Proceedings of the IEEE/CVF Conference on Computer Vision and Pattern Recognition (CVPR)},
  year      = {2025},
  pages     = {18011--18020}
}

@inproceedings{17,
  title     = {Timestep Embedding Tells: It's Time to Cache for Video Diffusion Model},
  author    = {Liu, Feng and Zhang, Shiwei and Wang, Xiaofeng and Wei, Yujie and Qiu, Haonan and Zhao, Yuzhong and Zhang, Yingya and Ye, Qixiang and Wan, Fang},
  booktitle = {Proceedings of the IEEE/CVF Conference on Computer Vision and Pattern Recognition (CVPR)},
  year      = {2025},
  pages     = {7353--7363}
}

@inproceedings{18,
  title     = {BlockDance: Reuse Structurally Similar Spatio-Temporal Features to Accelerate Diffusion Transformers},
  author    = {Zhang, Hui and Gao, Tingwei and Shao, Jie and Wu, Zuxuan},
  booktitle = {Proceedings of the IEEE/CVF Conference on Computer Vision and Pattern Recognition (CVPR)},
  year      = {2025},
  pages     = {12891--12900}
}

@inproceedings{19,
  title     = {CacheQuant: Comprehensively Accelerated Diffusion Models},
  author    = {Liu, Xuewen and Li, Zhikai and Gu, Qingyi},
  booktitle = {Proceedings of the IEEE/CVF Conference on Computer Vision and Pattern Recognition (CVPR)},
  year      = {2025},
  pages     = {23269--23280}
}

@inproceedings{20,
  title     = {Accelerating Diffusion Transformers with Token-wise Feature Caching},
  author    = {Zou, Chang and Liu, Xuyang and Liu, Ting and Huang, Siteng and Zhang, Linfeng},
  booktitle = {International Conference on Learning Representations (ICLR)},
  year      = {2025}
}

@inproceedings{21,
  title     = {From Reusing to Forecasting: Accelerating Diffusion Models with TaylorSeers},
  author    = {Liu, Jiacheng and Zou, Chang and Lyu, Yuanhuiyi and Chen, Junjie and Zhang, Linfeng},
  booktitle = {Proceedings of the IEEE/CVF International Conference on Computer Vision (ICCV)},
  year      = {2025},
  pages     = {15853--15863}
}

@inproceedings{22,
  title     = {OmniCache: A Trajectory-Oriented Global Perspective on Training-Free Cache Reuse for Diffusion Transformer Models},
  author    = {Chu, Huanpeng and Wu, Wei and Feng, Guanyu and Zhang, Yutao},
  booktitle = {Proceedings of the IEEE/CVF International Conference on Computer Vision (ICCV)},
  year      = {2025},
  pages     = {16302--16312}
}

@inproceedings{23,
  title     = {Accelerating Diffusion Transformer via Gradient-Optimized Cache},
  author    = {Qiu, Junxiang and Liu, Lin and Wang, Shuo and Lu, Jinda and Chen, Kezhou and Hao, Yanbin},
  booktitle = {Proceedings of the IEEE/CVF International Conference on Computer Vision (ICCV)},
  year      = {2025},
  pages     = {17608--17617}
}

@inproceedings{26,
  title     = {MosaicDiff: Training-free Structural Pruning for Diffusion Model Acceleration Reflecting Pretraining Dynamics},
  author    = {Guo, Bowei and Tang, Shengkun and Zeng, Cong and Shen, Zhiqiang},
  booktitle = {Proceedings of the IEEE/CVF International Conference on Computer Vision (ICCV)},
  year      = {2025},
  pages     = {1655--1664}
}

@inproceedings{27,
  title     = {Token Merging: Your ViT But Faster},
  author    = {Bolya, Daniel and Fu, Cheng-Yang and Dai, Xiaoliang and Zhang, Peizhao and Feichtenhofer, Christoph and Hoffman, Judy},
  booktitle = {International Conference on Learning Representations (ICLR)},
  year      = {2023}
}

@inproceedings{28,
  title     = {Token Merging for Fast Stable Diffusion},
  author    = {Bolya, Daniel and Hoffman, Judy},
  booktitle = {Proceedings of the IEEE/CVF Conference on Computer Vision and Pattern Recognition Workshops (CVPRW)},
  year      = {2023},
    pages     = {4599--4603}
}

@inproceedings{29,
  title     = {Importance-Based Token Merging for Efficient Image and Video Generation},
  author    = {Wu, Haoyu and Xu, Jingyi and Le, Hieu and Samaras, Dimitris},
  booktitle = {Proceedings of the IEEE/CVF International Conference on Computer Vision (ICCV)},
  year      = {2025},
  pages     = {4983--4995}
}

@inproceedings{30,
  title     = {SpargeAttention: Accurate and Training-free Sparse Attention Accelerating Any Model Inference},
  author    = {Zhang, Jintao and Xiang, Chendong and Huang, Haofeng and Wei, Jia and Xi, Haocheng and Zhu, Jun and Chen, Jianfei},
  booktitle = {Proceedings of the 42nd International Conference on Machine Learning (ICML)},
  year      = {2025},
  series    = {Proceedings of Machine Learning Research},
  volume    = {267},
  pages     = {76397--76413}
}

@article{31,
  title   = {Region-Adaptive Sampling for Diffusion Transformers},
  author  = {Liu, Ziming and Yang, Yifan and Zhang, Chengruidong and Zhang, Yiqi and Qiu, Lili and You, Yang and Yang, Yuqing},
  journal = {arXiv preprint arXiv:2502.10389},
  year    = {2025}
}

@article{32,
  title   = {SDiT: Semantic Region-Adaptive for Diffusion Transformers},
  author  = {Lin, Bowen and Ye, Fanjiang and Liu, Yihua and Guo, Zhenghui and Zhang, Boyuan and Zheng, Weijian and Xu, Yufan and Xing, Tiancheng and Wang, Yuke and Zhang, Chengming},
  journal = {arXiv preprint arXiv:2601.12283},
  year    = {2026}
}

@inproceedings{33,
  title     = {Pick-a-Pic: An Open Dataset of User Preferences for Text-to-Image Generation},
  author    = {Kirstain, Yuval and Polyak, Adam and Singer, Uriel and Matiana, Shahbuland and Penna, Joe and Levy, Omer},
  booktitle = {Advances in Neural Information Processing Systems (NeurIPS)},
  year      = {2023},
  pages = {36652--36663}
}

@inproceedings{34,
  title     = {ImageReward: Learning and Evaluating Human Preferences for Text-to-Image Generation},
  author    = {Xu, Jiazheng and Liu, Xiao and Wu, Yuchen and Tong, Yuxuan and Li, Qinkai and Ding, Ming and Tang, Jie and Dong, Yuxiao},
  booktitle = {Advances in Neural Information Processing Systems (NeurIPS)},
  year      = {2023},
  pages = {15903--15935}
}

@inproceedings{35,
  title     = {Rethinking the Spatial Inconsistency in Classifier-Free Diffusion Guidance},
  author    = {Shen, Dazhong and Song, Guanglu and Xue, Zeyue and Wang, Fu-Yun and Liu, Yu},
  booktitle = {Proceedings of the IEEE/CVF Conference on Computer Vision and Pattern Recognition (CVPR)},
  year      = {2024},
  pages = {9370--9379}
}

@inproceedings{36,
  title     = {Aesthetic Post-Training Diffusion Models from Generic Preferences with Step-by-step Preference Optimization},
  author    = {Liang, Zhanhao and Yuan, Yuhui and Gu, Shuyang and Chen, Bohan and Hang, Tiankai and Cheng, Mingxi and Li, Ji and Zheng, Liang},
  booktitle = {Proceedings of the IEEE/CVF Conference on Computer Vision and Pattern Recognition (CVPR)},
  year      = {2025},
  pages     = {13199--13208}
}

@inproceedings{37,
  title     = {AttriCtrl: A Generalizable Framework for Controlling Semantic Attribute Intensity in Diffusion Models},
  author    = {Chen, Die and Duan, Zhongjie and Li, Zhiwen and Chen, Cen and Chen, Daoyuan and Li, Yaliang and Chen, Yinda},
  booktitle = {International Conference on Learning Representations (ICLR)},
  year      = {2026}
}

@article{38,
  title   = {Playground v2.5: Three Insights towards Enhancing Aesthetic Quality in Text-to-Image Generation},
  author  = {Li, Daiqing and Kamko, Aleks and Akhgari, Ehsan and Sabet, Ali and Xu, Linmiao and Doshi, Suhail},
  journal = {arXiv preprint arXiv:2402.17245},
  year    = {2024}
}

@article{39, title="A Review of Artistic Image Generation and Evaluation: Models, Metrics, and Applications", author="Hanhoon Kang and Sung-Dong Moon", journal="KSII Transactions on Internet and Information Systems",  volume={20}, number={1}, year="2026", pages={23-37}}

@inproceedings{40,
  title     = {Dynamic Prompt Optimizing for Text-to-Image Generation},
  author    = {Mo, Wenyi and Zhang, Tianyu and Bai, Yalong and Su, Bing and Wen, Ji-Rong and Yang, Qing},
  booktitle = {Proceedings of the IEEE/CVF Conference on Computer Vision and Pattern Recognition (CVPR)},
  year      = {2024},
  pages     = {26627--26636}
}

@article{41,
  title   = {Human Preference Score v2: A Solid Benchmark for Evaluating Human Preferences of Text-to-Image Synthesis},
  author  = {Wu, Xiaoshi and Hao, Yiming and Sun, Keqiang and Chen, Yixiong and Zhu, Feng and Zhao, Rui and Li, Hongsheng},
  journal = {arXiv preprint arXiv:2306.09341},
  year    = {2023}
}

@inproceedings{42,
  title     = {LAION-5B: An Open Large-Scale Dataset for Training Next Generation Image-Text Models},
  author    = {Schuhmann, Christoph and Beaumont, Richard and Vencu, Romain and Gordon, Cade and Wightman, Ross and Cherti, Mehdi and Coombes, Theo and Katta, Aarush and Mullis, Clayton and Wortsman, Mitchell and others},
  booktitle = {Advances in Neural Information Processing Systems (NeurIPS)},
  year      = {2022},
  pages = {25278--25294}
}

@inproceedings{43,
  title     = {Scaling Rectified Flow Transformers for High-Resolution Image Synthesis},
  author    = {Esser, Patrick and Kulal, Sumith and Blattmann, Andreas and Entezari, Rahim and M{\"u}ller, Jonas and Saini, Harry and Levi, Yam and Lorenz, Dominik and Sauer, Axel and Boesel, Frederic and Podell, Dustin and Dockhorn, Tim and English, Zion and Rombach, Robin},
  booktitle = {Proceedings of the 41st International Conference on Machine Learning (ICML)},
  series    = {Proceedings of Machine Learning Research},
  volume    = {235},
  pages     = {12606--12633},
  year      = {2024},
  publisher = {PMLR}
}

@misc{44,
  author = {{Black Forest Labs}},
  title  = {{FLUX.1 [dev]}},
  year   = {2024},
  url    = {https://huggingface.co/black-forest-labs/FLUX.1-dev},
  note   = {Accessed 28 June 2026}
}

@inproceedings{100,
  title={Training language models to follow instructions with human feedback},
  author={Ouyang, Long and Wu, Jeffrey and Jiang, Xu and Almeida, Diogo and Wainwright, Carroll and Mishkin, Pamela and Zhang, Chong and Agarwal, Sandhini and Slama, Katarina and Ray, Alex and others},
  booktitle={Advances in Neural Information Processing Systems (NeurIPS)},
  year={2022},
  pages = {27730--27744}
}

@inproceedings{101,
  title={Deep reinforcement learning from human preferences},
  author = {Christiano, Paul F and Leike, Jan and Brown, Tom and Martic, Miljan and Legg, Shane and Amodei, Dario},
  booktitle = {Advances in Neural Information Processing Systems (NeurIPS)},
  year={2017}
}

@inproceedings{103,
  title     = {{GLIDE}: Towards Photorealistic Image Generation and Editing with Text-Guided Diffusion Models},
  author    = {Nichol, Alexander Quinn and Dhariwal, Prafulla and Ramesh, Aditya and Shyam, Pranav and Mishkin, Pamela and McGrew, Bob and Sutskever, Ilya and Chen, Mark},
  booktitle = {Proceedings of the 39th International Conference on Machine Learning (ICML)},
  series    = {Proceedings of Machine Learning Research},
  volume    = {162},
  pages     = {16784--16804},
  editor    = {Chaudhuri, Kamalika and Jegelka, Stefanie and Song, Le and Szepesvari, Csaba and Niu, Gang and Sabato, Sivan},
  year      = {2022},
  publisher = {PMLR}
}

@inproceedings{104,
  title={AVA: A Large-Scale Database for Aesthetic Visual Analysis},
  author={Murray, Naila and Marchesotti, Luca and Perronnin, Florent},
  booktitle={2012 IEEE Conference on Computer Vision and Pattern Recognition (CVPR)},
  year={2012}
}

@article{105,
  title   = {NIMA: Neural Image Assessment},
  author  = {Talebi, Hossein and Milanfar, Peyman},
  journal = {IEEE Transactions on Image Processing},
  year    = {2018}
}
